\documentclass[Afour,sageh,times]{ieeeconf}

\pdfoutput=1

\usepackage{mdwlist}

\usepackage{graphics} 
\usepackage{epsfig} 
\usepackage{mathptmx} 
\usepackage{amsmath} 
\usepackage[utf8]{inputenc}
\usepackage{graphicx}
\usepackage{tikz}
\usepackage{tabularx}
\usepackage{caption}
\usepackage{amsfonts}
\usepackage{amssymb}
\usepackage{bm}

\usepackage[noadjust]{cite}
\usepackage{epstopdf}
\usepackage{url}

\usepackage[colorlinks,bookmarksopen,bookmarksnumbered,citecolor=red,urlcolor=red]{hyperref}

\newcommand\BibTeX{{\rmfamily B\kern-.05em \textsc{i\kern-.025em b}\kern-.08em
T\kern-.1667em\lower.7ex\hbox{E}\kern-.125emX}}

\begin{document}

\title{3LP: a linear 3D-walking model including torso and swing dynamics}

\author{Salman Faraji$^{*}$ and Auke J. Ijspeert$^{*}$ \thanks{$^{*}$Biorobotics Laboratory, Ecole Polytechnique F\'ed\'erale de Lausanne (EPFL),  Lausanne, Switzerland}}

\maketitle

\begin{abstract}
In this paper, we present a new model of biped locomotion which is composed of three linear pendulums (one per leg and one for the whole upper body) to describe stance, swing and torso dynamics. In addition to double support, this model has different actuation possibilities in the swing hip and stance ankle which could be widely used to produce different walking gaits. Without the need for numerical time-integration, closed-form solutions help finding periodic gaits which could be simply scaled in certain dimensions to modulate the motion online. Thanks to linearity properties, the proposed model can provide a computationally fast platform for model predictive controllers to predict the future and consider meaningful inequality constraints to ensure feasibility of the motion. Such property is coming from describing dynamics with joint torques directly and therefore, reflecting hardware limitations more precisely, even in the very abstract high level template space. The proposed model produces human-like torque and ground reaction force profiles and thus, compared to point-mass models, it is more promising for precise control of humanoid robots. Despite being linear and lacking many other features of human walking like CoM excursion, knee flexion and ground clearance, we show that the proposed model can predict one of the main optimality trends in human walking, i.e. nonlinear speed-frequency relationship. In this paper, we mainly focus on describing the model and its capabilities, comparing it with human data and calculating optimal human gait variables. Setting up control problems and advanced biomechanical analysis still remain for future works. 
\end{abstract}

\section{Introduction}

Humanoid robots are challenging to control mainly due to their complex structure and floating base. During locomotion tasks, these systems introduce another complexity compared to wheeled or flying robots, which is the hybrid nature of stepping where the continuous model changes in each phase. Since the beginning, it has always been challenging to balance these robots considering the fact that they can only establish unilateral support with the environment. In addition, creating a sequence of motion, associated timing and the required control architecture in each phase of motion are other important topics in controlling humanoid robots. The main objectives are therefore being human-like, energy efficient, versatile and of course agile like humans. In this paper, we are proposing a new template model that describes main aspects of walking while being computationally very efficient. Such model can be very useful in modern control architectures from the computational perspective. It can also go beyond conventional template models such as Linear Inverted Pendulum (LIP) by producing more natural motions in faster walking speeds, resembling human locomotion. 

The design of controllers should address many concerns like fast implementation, stability, robustness to unknown model parameters and the degree of dependency on sensory data. It has always been appreciated if transition to different speeds and large disturbance rejection are handled in the same control framework. Therefore, candidate methods are normally coming with proper identification of the basin of attraction regarding system states, actuator limitations and violation of model assumptions. This identification is not always straight forward due to its nonlinear nature, though it has been postulated that two steps are enough to stabilize in almost all conditions \cite{zaytsev2015two}. In this regard, Model Predictive Control (MPC) is a powerful framework as it can find optimal policies constrained to certain actuation and state limitations. It can also predict if there is no feasible solution, in order to let the algorithm take a different decision. 

\subsection{Hierarchical controllers}
Recently, hierarchical control approaches are becoming popular, where a simple template model determines the overall dynamics in an abstract way and then, a detailed full-body inverse dynamics controller converts this behavior to individual actuator inputs \cite{faraji2014robust, feng20133d, kuindersma2014efficiently}. In dynamical systems, prediction of future evolution is mainly sensitive to the model and sensory data precision \cite{bhounsule2015discrete}. In hierarchical approaches similarly, dynamical matching between the template and the full model is crucial to ensure precise execution of the abstract plan. 

\begin{figure*}[th]
  \begin{tabular}{>{\arraybackslash}p{0.06\linewidth} >{\centering\arraybackslash}p{0.09\linewidth}  >{\centering\arraybackslash}p{0.09\linewidth} >{\centering\arraybackslash}p{0.11\linewidth} >{\centering\arraybackslash}p{0.10\linewidth} >{\centering\arraybackslash}p{0.10\linewidth} >{\centering\arraybackslash}p{0.13\linewidth} >{\centering\arraybackslash}p{0.13\linewidth}}
    & \includegraphics[page=10,trim = 50mm 20mm 50mm 95mm, clip, width=0.10\textwidth]{figures/newmodel.pdf} & 
    \includegraphics[page=1,trim = 40mm 20mm 50mm 95mm, clip, width=0.10\textwidth]{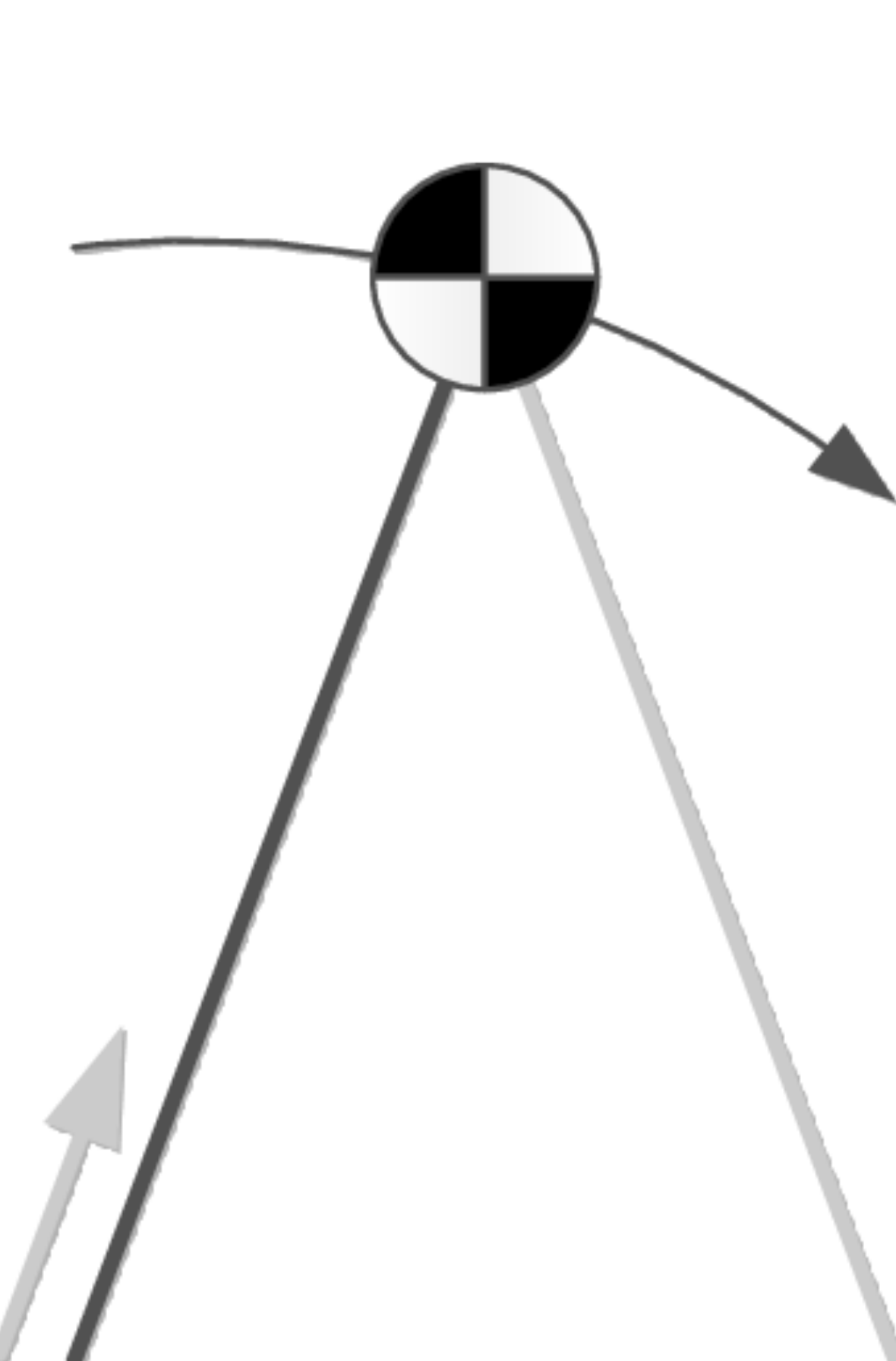} & 
    \includegraphics[page=2,trim = 35mm 20mm 35mm 95mm, clip, width=0.12\textwidth]{figures/newmodel.pdf} & 
	\includegraphics[page=3,trim = 40mm 20mm 35mm 95mm, clip, width=0.11\textwidth]{figures/newmodel.pdf} & 
    \includegraphics[page=4,trim = 35mm 20mm 45mm 95mm, clip, width=0.11\textwidth]{figures/newmodel.pdf} & 
    \includegraphics[page=8,trim = 15mm 35mm 20mm 80mm, clip, width=0.14\textwidth]{figures/newmodel.pdf} & 
    \includegraphics[page=13,trim = 35mm 20mm 20mm 95mm, clip, width=0.13\textwidth]{figures/newmodel.pdf} \\
    & \cite{iida2009toward} & \cite{hemami1977inverted} & \cite{asano2004novel} & \cite{byl2008approximate} & \cite{asano2004novel} & \cite{kuo1999stabilization} & \cite{kajita1991study} \\ 
    name & SLIP & IP & - & - & - & - & LIP \\
    knee & - & - & - & - & x & - & - \\
    steering & - & - & - & - & - & - & x \\
    3D & - & - & - & - & - & x & x \\
    smooth & x & - & - & - & - & - & x \\
    torso & - & - & - & - & - & - & - \\
    swing & - & - & x & x & x & x & - \\
    linear & - & - & - & - & - & - & x \\
    \hline \\
    & \includegraphics[page=11,trim = 50mm 30mm 50mm 35mm, clip, width=0.10\textwidth]{figures/newmodel.pdf} &
    \includegraphics[page=12,trim = 50mm 20mm 50mm 35mm, clip, width=0.10\textwidth]{figures/newmodel.pdf} & 
    \includegraphics[page=5,trim = 35mm 20mm 35mm 35mm, clip, width=0.12\textwidth]{figures/newmodel.pdf} &
    \includegraphics[page=6,trim = 50mm 20mm 35mm 35mm, clip, width=0.11\textwidth]{figures/newmodel.pdf} &
    \includegraphics[page=7,trim = 30mm 20mm 45mm 35mm, clip, width=0.11\textwidth]{figures/newmodel.pdf} &
    \includegraphics[page=9,trim = 20mm 20mm 20mm 20mm, clip, width=0.14\textwidth]{figures/newmodel.pdf} &
    \includegraphics[page=14,trim = 30mm 20mm 20mm 35mm, clip, width=0.14\textwidth]{figures/newmodel.pdf} \\
    & \cite{maufroy2011simplified} & \cite{maziar2015FMHC} & \cite{hasaneini2013dynamic} & \cite{gomes2011walking} & \cite{manchester2014real} & \cite{gregg2012control} & proposed \\
    name & BSLIP & FMCH & - & - & - & - & 3LP \\
    knee & - & - & - & - & x & x & - \\
    steering & - & - & - & - & - & x & (x) \\
    3D & - & - & - & - & - & x & x \\
    smooth & x & x & - & x & - & - & x \\
    torso & rotating & rotating & rotating & rotating & rotating & rotating & fixed  \\
    swing & - & - & x & x & x & x & x \\
    linear & - & - & - & - & - & - & x \\
    \hline
  \end{tabular}
  \caption{Different key models introduced in literature for walking. In this table, the model is standing on the left leg and the right leg is in swing motion. Solid arrows show the direction of motion and degrees of freedom while gray arrows show actuations torques or push-off forces. For models without swing dynamics, we show the swing leg also in gray color to implicitly show that attack angle is a control authority. Note that some of these models are only in 2D while more advanced models are in 3D. Some models simulate pelvis width, torso dynamics or ground clearance as well. Most of these models produce compass gait, however some have ankle actuation or arc foot. In the comparison table, we mention important features such as knee flexion (for ground clearance), steering capabilities, 3D model, smooth profiles, inclusion of torso and swing dynamics and providing linear equations. By smoothness we mean no collision and push-off impulse, but possibly describing double support phase. Many of these models allow for torso pitch while we keep it fixed in 3LP for simplicity. Note that 3LP can describe steering like our previous work with LIP \cite{faraji2014robust} only if pelvis width is set to zero. Overall, 3LP offers many features that are not existing in other template models, although it is still linear like LIP. }
  \label{fig::models}
\end{figure*}

\subsection{Inverted Pendulums}
One of the earliest template models that roughly described human in single support was Inverted Pendulum (IP) \cite{mcgeer1990passive} with fixed leg length. In this model, a single mass rolls over a contact point, established by a massless leg. IP is widely used to analyze passive walkers \cite{mcgeer1990passive} and human motion \cite{kuo2005energetic}. Inspired by this idea, there are various simple robots built to walk naturally with minimal energy, pumped either in push-off or swing hip or both \cite{collins2005efficient}. Later this model was simplified to Linear Inverted Pendulum (LIP) \cite{kajita2003biped}, mainly favoring availability of analytical solutions instead of numerical integration. With proper modulation of Zero Moment Point (ZMP) \cite{zmp_full_review}, many position controlled robots like ASIMO \cite{sakagami2002intelligent} can perform walking through inverse kinematics methods. These algorithms are normally able to produce slow to moderate walking speeds. However complex robots using the LIP method usually walk with crouched knees to keep the Center of Mass (CoM) at constant height. In addition to increasing energy consumption, it is harmful for the robot in long term and less human like, though providing full controllability. 

\subsection{Multi-link pendulums}
Apart from these two models, there are other nonlinear extensions solved numerically. In \cite{byl2008approximate, asano2004novel} the IP model is extended to have two separate masses for each leg as well as a single mass at hip. Using similar actuation schemes, this model produces compass gaits on 2D-constrained simple robots. In \cite{asano2004novel}, same model is modified to have another Degree of Freedom (DoF) in the knee for the swing leg to avoid foot scuffing. The stance leg however always stays straight. In \cite{westervelt2007feedback}, this advanced model is augmented with a torso and later, it is used also by \cite{manchester2014real} to perform natural walking on uneven terrain using a library of motion primitives. Another model with four masses in legs, hip and torso is proposed in \cite{gregg2009bringing} without any DoF in the knees, trying to slightly turn in 3D. 

Targeting impact-less walking, a simpler model with two passive springs in the hips is proposed by Gomes \cite{gomes2011walking}. Addition of these springs is mainly motivated by elastic properties of human muscles. By exploiting torso motions, Gomes can find zero energy gaits that means impact-losses are less important in energetics of walking. Another very interesting extension of IP-based models is proposed in \cite{gregg2012control}. In this new model, the pelvis has a width in 3D with a mass in the center. The swing leg is also having another DoF in the knee. This new 3D model is allowed to take advantage of a limited transversal wrench in the contact point for better turning. However finding a periodic gait for such complicated model is difficult and computationally expensive. 

\subsection{Spring-Loaded Pendulums}
It is always questionable which template model produces more realistic motion. This can be inspected from the viewpoint of geometry, torques or energy. The aforementioned categories mainly address energetic and geometric similarities. However ground reaction forces and elastic behaviors are other favorite aspects addressed by another category based on Spring Loaded Inverted Pendulum (SLIP). This model is composed of two massless springs (legs) connected to a single mass. One can expect better description of energy exchange in this model over faster walking speeds and running, mimicking compliant properties of human tendons. Based on this model, Iida \cite{iida2009toward} built a hip-actuated robot walking in 2D with various springs, similar to human muscles. Properties of the passive version without hip actuation were later widely explored in \cite{rummel2010stable}. Using the concept of Virtual Pivot Point (VPP) to stabilize the torso, the model was also extended to have an upper body \cite{maziar2015FMHC}. In Figure.\ref{fig::models}, we have briefly shown key models proposed for walking in the literature. Note that in some of them, passive springs are added to the hip actuators for energy storage, similar to humans. In this paper however, we do not investigate elastic behaviors and energy storage. 

\subsection{Control difficulty}
Except LIP, all other models presented before require numerical integration to obtain time trajectories. Therefore, the Jacobian around a nominal solution linearizes the model and provides the framework for Floquet analysis or discrete controller design \cite{rummel2010stable}. This approach can be used to create an optimal library of primitives \cite{kelly2015non, manchester2014real, gregg2012control}. However online reaction to disturbances as well as inclusion of other inequality constraints that are often ignored in calculating a stable basin of attraction, limit the generality of this framework. MPC on the other hand is powerful in this regard, however it requires simple and possibly linear models to provide online performance, depending on the hardware platform. LIP model therefore fits best in the MPC framework \cite{faraji2014robust, herdt2010walking}. MPC, its simpler version, LQR and sometimes Discretized LQR (DLQR) \cite{ogata1995discrete} are used a lot on nonlinear models to stabilize walking gaits and recover pushes \cite{kelly2015non, byl2008approximate}. Either a library of optimal policies is generated off-line or a discrete transition model is considered at specific events like CoM apex or heel-strike. 

\subsection{Why 3LP?}
In this paper, we propose a more general version of LIP \cite{kajita2003biped} with three linear pendulums (called 3LP) that captures torso and swing dynamics in 3D. This model allows prediction of future at any time in closed form which is favorable by limited computational resources and MPC. Compared to LIP, the planned motion with 3LP has instantaneous accelerations that are easier to track by inverse dynamics block in the hierarchical control. In other words, the motion of CoM is more natural for the humanoid robot as swing and torso dynamics are taken into account. Additionally, the swing trajectory is more natural compared to many other template models that track an imposed angle of attack with a stiff controller \cite{kelly2015non,byl2008approximate,collins2005efficient}. It should be noted however that the CoM height in our model will still be constant similar to LIP. 

The 3LP model supports many different inputs, i.e. hip and ankle actuation authorities. These input dimensions let us find various types of gaits with simple algebraic routines, not optimizations. 3LP as a template model is in fact very useful for motion planning. Besides describing falling dynamics like IP \cite{kuo2005energetic} and LIP \cite{kajita20013d}, hip torques required to keep the torso upright are also part of the model like \cite{maufroy2011simplified}. The most outstanding feature of 3LP is considering swing dynamics that allow us to calculate natural cycles for the model. Considering Figure.\ref{fig::models} again, there are few models in the literature that consider this integral part of walking. In these models due to nonlinearity, numerical integration is always needed to search for periodic gaits. In 3LP however we do not need to impose the timing, nor optimizing hip torque trajectories. 

This paper merely focuses on introducing 3LP and its capabilities, while setting up control problems remain for future work. The main motivation behind 3LP is for control however, as it can provide more natural CoM and swing motions. It suits MPC control better than LIP in our previous work \cite{faraji2014robust}, because we do not need to define a desired footstep plan. The plan comes out of 3LP just by modulating hip torques. Closed form formulations also let the MPC controller change phase timings in case of extreme push recovery for example. Although the MPC problem becomes nonlinear in this case, one can define meaningful hip torque magnitude and rate limitations instead of putting vague timing boundaries on the step time which do not precisely reflect physical facts about the real hardware.

In the next section, we will explain the model details and assumptions behind as well as compare it to the existing models. Next, a method to find different periodic gaits is explained based on symmetry ideas. In the end we show that for a human-like gait, actuation profiles of 3LP are similar to those of human. Additionally, we also show quantitatively that 3LP can explain main optimality criteria of human walking, i.e. speed-frequency relation, despite being linear. 

\section{3LP dynamics}
\label{sec::model}

\begin{figure}[]
        \centering
        \includegraphics[trim = 10mm 0mm 0mm 3mm, clip, width=0.4\textwidth]{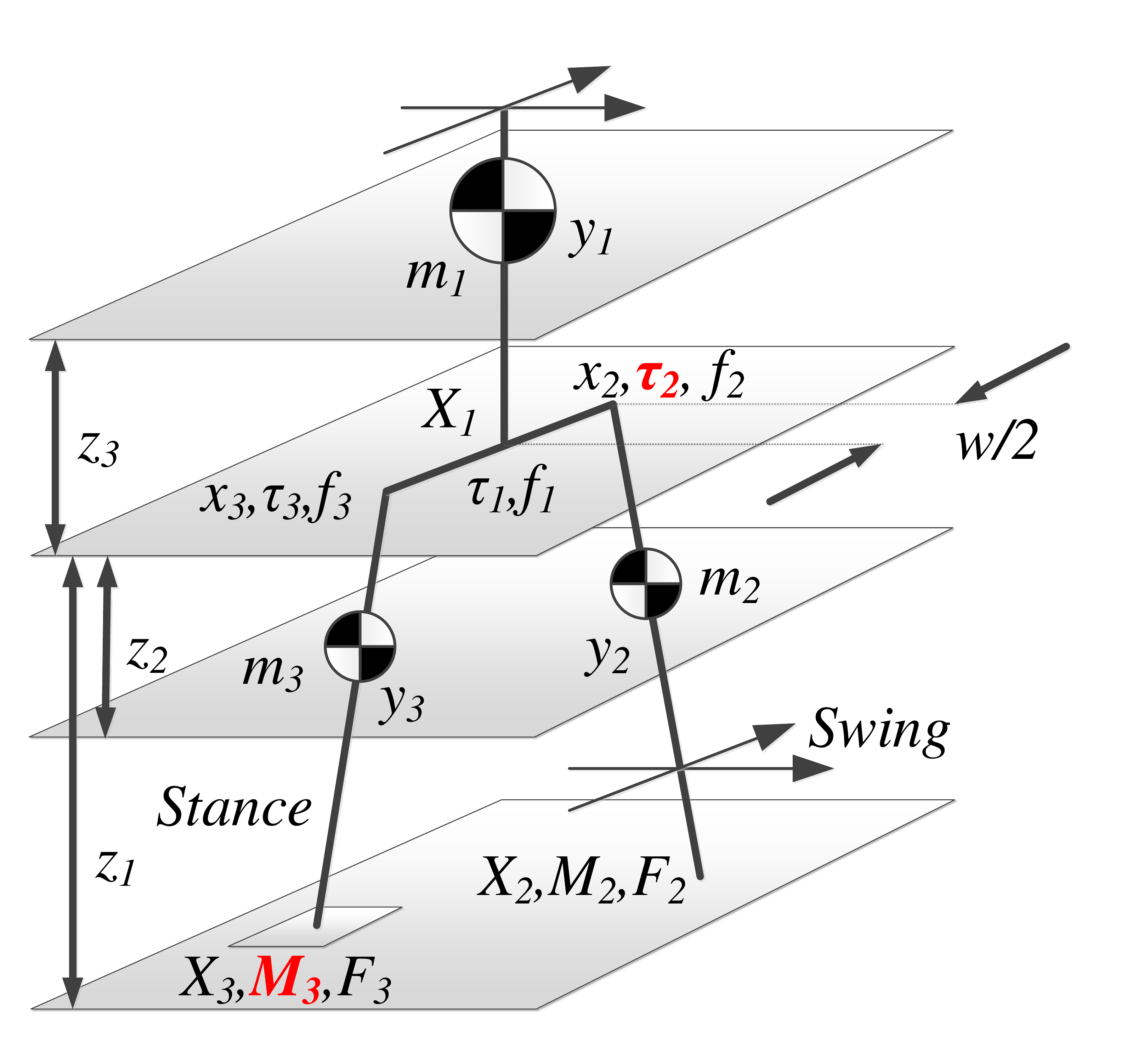}
        \caption{A schematic of 3LP model with all variables and parameters. The bottom plane shows level ground and all upper planes of fixed height show where the three masses and the pelvis are constrained to move. The torso is always upright and the pelvis is along y axis, by model construction. The swing foot remains inside the bottom plane, i.e. sliding on the ground with zero force during swing phase. Note that in single support all contact forces for the swing leg ($F_2,M_2$) are zero. The arbitrary actuation inputs are $\tau_2$ and $M_3$ shown in red and bold.  } 
        \label{fig::model}
\end{figure}

Motivated by the fact that CoM motion is influenced by swing and torso dynamics, we have added two other pendulums to the normal linear inverted pendulum model, connected together with a pelvis of certain width. In this model, as shown in Figure.\ref{fig::model}, there are 2-DoF actuators in each hip and ankle. We demonstrate feet of limited size in Figure.\ref{fig::model} merely to mention availability of ankle actuation. The upper body is represented by a single mass referred to as torso and each leg is represented by a single inertia-less mass. By construction (assuming ideal controllers), masses stay in horizontal planes of constant height and the torso is always upright without transversal rotation. These assumptions are used in \cite{gregg2012control} as well to decouple sagittal and lateral dynamics. Since the torso is connected to an accelerated frame (i.e. pelvis), the $\tau1$ torque is always found to keep the torso upright. Such hip torques in literature are alternatively calculated by virtual pendulum concept in the models which let the torso pitch or roll freely during the motion \cite{maziar2015FMHC}. Inspired by the fact that rolling contact constraint produces more human like walking \cite{hamner2013rolling}, we also allow for transversal wrenches at stance contact to keep the pelvis orientation fixed. In 3LP model, we do not consider turning properties as they make the model nonlinear. It is however practically easy to turn the robot using inverse dynamics layer, as we showed in \cite{faraji2014robust} using a simple LIP model. There is no need to add heel-strike and push-off impulses since our double support phase smoothly takes care of transition. In SLIP based models \cite{maziar2015FMHC}, the compliant springs automatically produce a double support phase and perform weight transition without impulses. In 3LP however, switching to double support is triggered when both horizontal components of the swing foot velocity are zero. This assumption is typically used in models with swing dynamics and double support phase like \cite{gomes2011walking}, though unlike 3LP, Gomes removes double support for more simplicity.   

In Figure.\ref{fig::model}, all external/internal forces and torques as well as positions are shown for each interesting point of the model, i.e. contacts, hips and pelvis center. These vector variables in our model are expressed in Cartesian frame. One can easily write geometric relations as:
\begin{eqnarray}
	\nonumber X_{1} &=& [X_{1,x},X_{1,y},z_{1}]^T \\
	\nonumber X_{2} &=& [X_{2,x},X_{2,y},0]^T \\
	\nonumber X_{3} &=& [X_{3,x},X_{3,y},0]^T \\
	\nonumber x_{2} &=& X_{1} + [0,\frac{wd}{2},0]^T \\
	\nonumber x_{3} &=& X_{1} + [0,-\frac{wd}{2},0]^T \\
	\nonumber y_{1} &=& X_{1} + [0,0,z_{3}]^T \\
	\nonumber y_{2} &=& x_{2} + \frac{z_{2}}{z_{1}} (X_{2}-x_{2}) \\
	y_{3} &=& x_{3} + \frac{z_{2}}{z_{1}} (X_{3}-x_{3})
\end{eqnarray}
Where $w$ stands for pelvis width and $d =\pm 1$, depending on left or right support. We can write total force equations for each mass i=1,2,3:
\begin{eqnarray}
	m_{i} (\ddot{y}_{i} + g) = f_{i} + F_{i} 
\end{eqnarray}
and total moment as:
\begin{eqnarray}
	\nonumber (X_{1}-y_{1})\times f_{1} + M_{1} + \tau_{1} = 0 \\
	\nonumber (X_{2}-y_{2})\times F_{2} + (x_{2}-y_{2})\times f_{2} + M_{2} + \tau_{2} = 0 \\
	(X_{3}-y_{3})\times F_{3} + (x_{3}-y_{3})\times f_{3} + M_{3} + \tau_{3} = 0 
\end{eqnarray}
Finally, we can write these equations for our mass-less pelvis around the center point:
\begin{eqnarray}
	\nonumber -f_{1} -f_{2} -f_{3} = 0 \\
	-\tau_{1} -\tau_{2} -\tau_{3} - \frac{wd}{2} (f_{2}-f_{3})=0
\end{eqnarray} 
which link all other variables together. In these equations, we consider $X_{1},X_{2},X_{3}$ as independent variables and solve for others as dependent variables. Actuation possibilities for 3LP are selected as stance foot $M_{3}$ and swing hip torques $\tau_{2}$ in sagittal and lateral directions. Note that the variables $F_{1},M_{1}$ stand for disturbances added to our model to simulate external forces. Now, the differential equations governing the system behavior could be obtained for different phases. Features of a full stride, consisting of a double support followed by a single support are defined in Table.\ref{table::phase}. In double support, the weight is transfered from $F_2$ to $F_3$ and in single support, the leg with variables of subscripts 2 will swing forward. We are going to find transfer matrices that relate the states of the system together at any instance of time over a stride phase.

\begin{table}[htbp]
\centering
  \begin{tabular}{ccc}
  \hline
  Full stride =  & double support + & single support \\ \hline
  duration & $T_{ds}$ & $T_{ss}$ \\
  timing order & 1 & 2 \\ 
  stance leg & subscripts 3 & subscripts 3 \\
  swing leg & X & subscripts 2 \\
  control input & ankle & hip/ankle \\
  controllability & redundant & full \\
  \hline
  \end{tabular}
  \caption{Information about the two consecutive phases that form a full stride phase.}
  \label{table::phase}
\end{table}

\subsection{Single support}
In this phase, the swing foot does not have any external forces, meaning:
\begin{eqnarray}
	\nonumber F_{2}=[0,0,0]^T \\ M_{2}=[0,0,0]^T
	\label{eqn::zeroforces}
\end{eqnarray}
Also stance foot is fixed on the ground, meaning $X_{3}=const$. The position variable $X$, disturbance vector $W$ and contact position $P$ are defined as:
\begin{eqnarray}
	X = \begin{bmatrix} X_{2,x} \\ X_{2,y} \\ X_{1,x} \\ X_{1,y} \end{bmatrix},
	W = \begin{bmatrix} F_{1,x} \\ F_{1,y} \\ M_{1,y} \\ M_{1,x} \end{bmatrix}, 
	P = \begin{bmatrix} X_{3,x} \\ X_{3,y} \end{bmatrix},
	\label{eqn::vars} 
\end{eqnarray}
And the inputs to the system are defined as:
\begin{eqnarray}
	\begin{bmatrix} \tau_{2,y} \\ \tau_{2,x} \\ M_{3,y} \\ M_{3,x} \end{bmatrix} = 
	\begin{bmatrix} M_{h,y} \\ M_{h,x} \\ M_{a,y} \\ M_{a,x} \end{bmatrix} + 
	\frac{t}{T_{ss}} \begin{bmatrix} rM_{h,y} \\ rM_{h,x} \\ rM_{a,y} \\ rM_{a,x} \end{bmatrix} = U + \frac{t}{T_{ss}} rU
	\label{eqn::inputs}
\end{eqnarray}
Note that we consider two modes of input torques, constant and linearly increasing with time (ramp profile). Therefore we have eight inputs to the system. Now for single support phase, we write linear differential equations and using Maple software \cite{Maple10}, we solve them analytically to find the state evolution after arbitrary time with respect to the initial state.
\begin{eqnarray}
	\nonumber &\ddot{X} = C_1 X + C_2(t) \begin{bmatrix} P^T & U^T & rU^T & W^T & d^T \end{bmatrix}^T \\
	& Q(t) = H^{ss}(t) Q(0)
	\label{eqn::differentialss}
\end{eqnarray} 
where $C_1$ is a constant matrix and $C_2(t)$ is a linear function of time. Here the vector $Q$ contains both inputs and states together:
\begin{eqnarray}
	\nonumber & Q(t) = \\ &\begin{bmatrix} X(t)^T & \dot{X}(t)^T & P^T & U^T & rU^T & W^T & d^T \end{bmatrix}^T
	\label{eqn::full_vector}
\end{eqnarray}
Note that inputs $U$ and $rU$ are constant during a single support phase. As expected, $H^{ss}(t)$ could be written as:
\begin{eqnarray}
	H^{ss}(t) = H^{ss}_0 + \sum_{i=1}^{4} H^{ss}_i e^{w^{ss}_it} + H^{ss}_5t
	\label{eqn::Hss_decompose}
\end{eqnarray}
where $w^{ss}_i$ and $H^{ss}_i$ variables are complicated functions of system model parameters, though very sparse. Once these individual matrices are calculated off-line, $H^{ss}(t)$ can be easily calculated online by few arithmetic operations. It is also worth mentioning that lateral and sagittal motions are decoupled, meaning that in $H^{ss}(t)$, only elements with both odd or both even column and row index are nonzero. We also like to emphasize that the variables $P,U,rU,W,d$ are assumed to be constant during the whole single support phase and therefore, the corresponding rows on $H^{ss}(t)$ only have $1$ on the diagonal. We keep everything packed for simpler manipulation of matrices in future. Note that in $\dot{X}(0)$, initial foot velocities are zero, starting to swing from rest conditions. It should also be noted that the switch between left and right support only changes the variable $d$ and contact positions $P$, not the system transition matrix $H^{ss}(t)$.

\subsection{Double support}
In this phase, the two feet are fixed and contact forces are being transfered from $(F_{2},M_{2})$ to $(F_{3},M_{3})$. Note that although there are two constraints added regarding the new fixed foot, 6 other forces and torques $(F_{2},M_{2})$ are none-zero and should be found. Therefore we need 8 additional equations to replace (\ref{eqn::zeroforces}) and (\ref{eqn::inputs}). In our double support, we decide to linearly transfer the weight from one leg to another. This means the vertical component of the Ground Reaction Force (GRF) in previous stance leg (which is constant during single support) will go linearly to zero during double support time. Such policy will ensure smoother torque profiles which are easier to track on the real robot. The linear policy also makes simpler equations in analytic form (compared to quadratic or other forms). 

Note that all contact forces are calculable at any time including right at the end of single support (as a function of state variables). If we want to preserve continuity of other horizontal components of GRF in phase transitions as well, in the right hand side of the differential equations (\ref{eqn::differentialss}), terms like $t x(t)$ will appear. Although these forms are still linear, they make finding analytical solutions difficult. Instead, we keep the Center of Pressure (CoP) for each foot constant during double support. Note that in single support, ground reaction torques in the stance foot are determined by (\ref{eqn::inputs}) and the vertical GRF is constant. The CoP position can be then simply preserved in double support by linearly decreasing contact reaction moment $M_2$ in stance leg (and increasing $M_3$ accordingly). Overall, equations being considered additionally for the CoP constraints are:
\begin{eqnarray}
	\begin{bmatrix} M_{2,y} \\ M_{2,x} \end{bmatrix} &=& (1-\frac{t}{T_{ds}}) \begin{bmatrix} M_{a,y}+rM_{a,y} \\ -M_{a,x}-rM_{a,x} \end{bmatrix} \\
	\begin{bmatrix} M_{3,y} \\ M_{3,x} \end{bmatrix} &=& \frac{t}{T_{ds}} \begin{bmatrix} M_{a,y} \\ M_{a,x} \end{bmatrix}
\end{eqnarray}
Linear transfer of weight and transversal contact torques as well as continuity of hip torque profiles require 4 more equations. We solve all equations except 4, and replace variables in these 4 equations. This results in a set of 4 equations $E$ with 8 unknown variables, 6 hip torques and 2 vertical components of GRFs. Inspired by linear weight transfer idea, we obtain the following other 4 equations in (\ref{eqn::ds}), which transfer torques and forces uniformly during the double support:
\begin{eqnarray} []
	\nonumber V_2 = \begin{bmatrix}\tau_{2}^T, F_{2,z}\end{bmatrix}^T \\
	\nonumber V_3 = \begin{bmatrix}\tau_{3}^T, F_{3,z}\end{bmatrix}^T \\
	\nonumber \hat{V}_2 = \begin{bmatrix}M_{h,x}, & M_{h,y}, & 0, & 0\end{bmatrix}^T \\ 
	\nonumber \hat{V}_3 = \begin{bmatrix}-M_{h,x}-rM_{h,x}, & M_{h,y}+rM_{h,y}, & 0, & 0\end{bmatrix}^T \\ 
	\frac{1}{1-\frac{t}{Tds}} \frac{\partial E}{\partial V_2} (V_2-\hat{V}_2) =
	\frac{1}{\frac{t}{Tds}} \frac{\partial E}{\partial V_3} (V_3-\hat{V}_3)
	\label{eqn::ds}
\end{eqnarray}
Here we calculate the Jacobian of these equations with respect to $V_2$ and $V_3$, multiply them by these variables again and divide each side by proper time to induce linear force transition. Combining the last equation set in (\ref{eqn::ds}) with $E$ (to have 8 equations for 8 variables) and general equations of the robot mentioned before, similar to single support case, Maple finds analytical solutions of the form:
\begin{eqnarray}
	Q(t) = H^{ds}(t) Q(0)
	\label{eqn::differentialds}
\end{eqnarray} 
Note that variables $X_{2,x}$ and $X_{2,y}$ (foot positions) are constant during double support phase and therefore, they are correctly linked to the right-hand side vector in this equation. Again as before:
\begin{eqnarray}
	H^{ds}(t) = H^{ds}_0 + \sum_{i=1}^{4} H^{ds}_i e^{w^{ds}_it} + H^{ds}_5t
	\label{eqn::Hds_decompose}
\end{eqnarray}
Which ensures fast implementation of this matrix. At this stage, we would also like to define row selection matrices that if multiplied from left, they output corresponding rows. If their transpose is multiplied from right, they output corresponding columns alternatively, if dimensions match. These selection matrices are listed in Table.\ref{table::selectoin}.

\begin{table}[htbp]
\centering
  \begin{tabular}{cc}
  \hline
  Matrix & selected variables \\
  \hline
  $S_{XP} \in \mathbb{R}^{8 \times 23}$ & all states and $P$ except $\dot{X}_{2}$ \\
  $S_{\dot{X}_{2}} \in \mathbb{R}^{2 \times 23}$ & $\dot{X}_{2}$ \\
  $S_{X_{2,x}} \in \mathbb{R}^{1 \times 23}$ & sagittal component of $X_{2}$ \\
  $S_{U} \in \mathbb{R}^{8 \times 23}$ & all inputs \\
  $S_{M_h} \in \mathbb{R}^{2 \times 23}$ & constant hip torques\\
  $S_{M_a} \in \mathbb{R}^{2 \times 23}$ & constant contact torques \\
  $S_{rM_a} \in \mathbb{R}^{2 \times 23}$ & time-increasing contact torques \\
  $S_{d} \in \mathbb{R}^{1 \times 23}$ & $d$ \\
  \hline
  \end{tabular}
  \caption{Different selection matrices used hereafter.}
  \label{table::selectoin}
\end{table}

\subsection{Full stride}
\begin{figure}[]
        \centering
        \includegraphics[trim = 0mm 10mm 0mm 10mm, clip, width=0.5\textwidth]{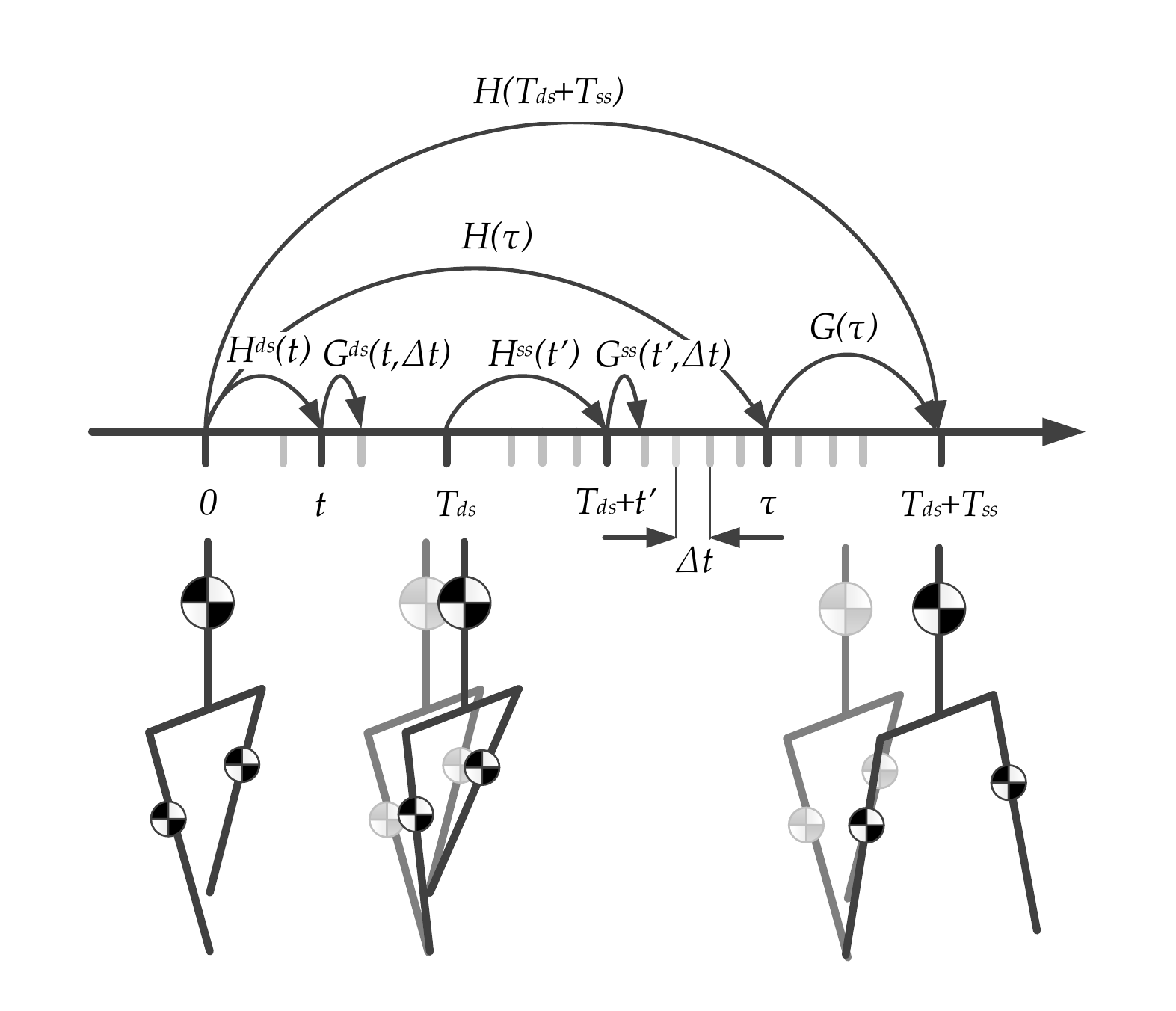}
        \caption{Definition of all transition matrices. Note that in this figure, $\Delta t$ represents arbitrary time duration that could be as small as the time-step for example, although there is no need to integrate the system. We use these matrices only for simulating intermittent pushes (discussed later) or for visualization. Otherwise the matrix $H(T_{stride})$ is enough to find the state transition.} 
        \label{fig::timing}
\end{figure}

Once we have matrices for both phases, we can link them together to find the transfer matrix for the full stride phase:
\begin{eqnarray}
	\nonumber H(t) = \left\{
	    \begin{array}{ll}
	        H^{ds}(t) & t \le T_{ds} \\
	        H^{ss}(t-T_{ds}) H^{ds}(T_{ds}) & 0 < t-T_{ds} \le T_{ss}
	    \end{array}
	\right.
\end{eqnarray}
Note that although we have used parameters $T_{ss}$ and $T_{ds}$ in calculating transfer matrices, they are defined by other methods explained later. The variable $T_{ds}$ is crucial for double support calculations, though $T_{ss}$ only determines the rate of time-increasing input components in single support. The duration of a stride phase is therefore defined as $T_{stride}=T_{ds}+T_{ss}$. Apart from the forward transfer matrix $H(t)$, we can also define another back transfer matrix in a similar way:
\begin{eqnarray}
	\nonumber H^{ss}(t+\Delta t) = G^{ss}(t,\Delta t) H^{ss}(t)  \\ 
	H^{ds}(t+\Delta t) = G^{ds}(t,\Delta t) H^{ds}(t)
\end{eqnarray}  
One can easily show that the matrices $G^{ss}$ and $G^{ds}$ could be written as:
\begin{eqnarray}
	\nonumber G^{ss}(t,\Delta t) = &(\sum_{i=0}^{4} ~^1\!G^{ss}_i e^{w^{ss}_i \Delta t} + ~^1\!G^{ss}_5 \Delta t) \\
	\nonumber + & t (\sum_{i=0}^{4} ~^2\!G^{ss}_i e^{w^{ss}_i \Delta t} + ~^2\!G^{ss}_5 \Delta t) \\
	\nonumber G^{ds}(t,\Delta t) = &(\sum_{i=0}^{4} ~^1\!G^{ds}_i e^{w^{ds}_i \Delta t} + ~^1\!G^{ds}_5 \Delta t) \\
	+ & t (\sum_{i=0}^{4} ~^2\!G^{ds}_i e^{w^{ds}_i \Delta t} + ~^2\!G^{ds}_5 \Delta t)
	\label{eqn::numerical}
\end{eqnarray}
which favor fast implementation. Now having $G^{ss}$ and $G^{ds}$ matrices, we can define a back transfer matrix as:
\begin{eqnarray}
	\nonumber G(t) = \left\{
	    \begin{array}{ll}
	        G^{ss}(T_{ss}) G^{ds}(t,T_{ds}-t) & t \le T_{ds} \\
	        G^{ss}(t-T_{ds},T_{ss}+T_{ds}-t) & 0 < t-T_{ds} \le T_{ss}
	    \end{array}
	\right.
\end{eqnarray}
Which always satisfies:
\begin{eqnarray}
	G(\tau) H(\tau) = H(T_{stride})
	\label{eqn::Gmatrix}
\end{eqnarray}
This equation indicates that any intermediate state could be evolved to the end of the phase simply by multiplying the matrix $G(t)$. For the purpose of visualization or time integration, one can also use individual matrices $G^{ss}$ and $G^{ds}$ with sufficiently small $\Delta t$. Figure.\ref{fig::timing} gives a better understanding of all transition matrices introduced so far as well as the composition of individual single and double support phases to make the full stride phase. Note that for an arbitrary state vector, it is straightforward to calculate all internal forces and torques in closed form as linear and quadratic functions of the state vector respectively. 

The matrix $H(t)$ is used to find evolution of the system state over time, specially at the end of the full stride where $t=T_{stride}$. Although the fact that the two feet are fixed in double support is already encoded in the matrix, one should consider another constraint for the system which forces the foot velocity to zero at the end of the stride. This means two input dimensions should be dedicated all the time and actively being regulated to ensure satisfaction of foot velocity constraint. For this purpose, we dedicate constant hip torque components for example:
\begin{eqnarray}
	H'(t) = H(t) - H(t)S_{M_h}^T(S_{\dot{X}_{2}}H(t)S_{M_h}^T)^{-1}S_{\dot{X}_{2}}H(t)
\end{eqnarray}
This equation in fact selects foot velocity rows, calculates the hip torques in terms of other variables and then replaces hip torques in other equations. Now using $H'(T_{stride})$, foot velocity is always zero at the end of the stride, though two input dimensions are lost as expected. One can do the same to find $G'(t)$ which is used to find the evolution of current state (measured at time $t$) until the end of the phase. In next sections we will use these new augmented matrices for simplicity.

So far in this section, we have found transition matrices for the system and defined a full phase, consisting of a double support followed by a single support. We have also formulated matrices such that there are very fast for online calculation. It should be noted that for the purpose of MPC, only a single off-line computation of the matrix $H(T_{stride})$ is enough. In next section, we are going to find different open-loop periodic gaits based on the type of actuation desired.

\section{Finding periodic gaits}
\label{sec::periodic}

All we need in this section is the matrix $H'(t)$ as the system is linear and fully expressed with this matrix. For the purpose of this paper unlike \cite{rummel2010stable}, we only focus on symmetric gaits observed in normal human walking. We find different classes of vectors that produce symmetric gaits. The concept of symmetry could be encoded in a single matrix along with the constraint of zero foot velocity at the end of the stride. Consider foot position $X_{2}$, pelvis position $X_{1}$, pelvis velocity $\dot{X}_{1}$ and stance foot position $P$ packed in a vector:
\begin{eqnarray}
	\begin{bmatrix} X_{2}^T & X_{1}^T & \dot{X_{1}}^T & P^T \end{bmatrix}^T
	\label{eqn::state_vector}
\end{eqnarray}
This vector can be of course extracted from the full vector $Q$ (\ref{eqn::full_vector}) with the selection matrix $S_{XP}$. After a stride, we define relative vectors between base, swing foot and stance foot which are linked to those in the beginning in a certain way. These vectors could be defined in the following matrix $M$, if multiplied by (\ref{eqn::state_vector}):
\begin{eqnarray}
\nonumber M = 
\begin{bmatrix} -1& . & 1 & . & . & . & . & . \\
				. &-1 & . & 1 & . & . & . & . \\
				. & . & 1 & . & . & . &-1 & . \\
				. & . & . & 1 & . & . & . &-1 \\
				. & . & . & . & 1 & . & . & . \\
				. & . & . & . & . & 1 & . & . \\
\end{bmatrix} 
\end{eqnarray}
Comparing these quantities before and after a symmetric stride, in sagittal plane, components are equal and in lateral plane, they are linked with a negative sign. This fact could be explained in a matrix $O$:
\begin{eqnarray}
	O = diag([1,-1,1,-1,1,-1])
\end{eqnarray}
Now we should also consider the transition matrix $T$ that exchanges the swing and support contact points after a stride as: 
\begin{eqnarray}
\nonumber T  = 
\begin{bmatrix} . & . & . & . & . & . & 1 & . \\
				. & . & . & . & . & . & . & 1 \\
				. & . & 1 & . & . & . & . & . \\
				. & . & . & 1 & . & . & . & . \\
				. & . & . & . & 1 & . & . & . \\
				. & . & . & . & . & 1 & . & . \\
				1 & . & . & . & . & . & . & . \\
				. & 1 & . & . & . & . & . & . \\
 \end{bmatrix} 
\end{eqnarray}
With these matrices, we can define the matrix $R$ which enfolds valid periodic gaits in its null-space:
\begin{eqnarray}
	R = -MS_{XP} + OMTS_{XP}H'(T_{stride})
	\label{eqn::periodic}
\end{eqnarray}
This matrix in fact compares aforementioned vector quantities before and after a stride. Note that apart from state vectors, the hip and contact torques $U$ and $Ur$, the disturbance vector $W$ and the variable $d$ could also be considered in calculating the null-space. However, it is meaningless in practice to consider a periodic gait with constant external disturbance. It should also be mentioned that for a valid solution vector, initial swing velocities are zero (impact-less model assumption) and contact positions $P$ are also set to zero to avoid redundant null-space dimensions. 

Remember that in the previous section, there were two variables to decide: $T_{ss}$ and $T_{ds}$. In this section, we find various types of gaits by selecting different combinations of actuation and timing. We do so by considering the matrix $R$ which is in fact a function of timing variables. Any periodic solution (a vector containing initial states and actuation inputs) should lie in the null-space of $R$ matrix. Note that only columns attributed to non-zero values in the solution vector are selected. We basically exclude columns related to initial foot velocity, contact position and disturbances to obtain the reduced matrix $R_0 \in \mathbb{R}^{8\times 15}$. We then find solution manifolds by combining different actuation dimensions. Indeed, possible actuations are swing hip and stance contact torques (in sagittal and lateral directions) and also their time-increasing modes. In the rest of this paper, we use human-like body parameters for numerical simulations, where mass distributions and geometries are taken from \cite{de1996adjustments}. Table\ref{table::params} lists these parameters for two adult-size and kid-size models, used further in this paper.
\begin{table}[htbp]
  \centering
  \begin{tabular}{cccc}
  \hline
  Model  & adult-size & kid-size & unit \\ \hline
  Total mass  & 70 & 30 &  kg\\
  Body length  & 1.7 & 1.0 & m  \\
  z1  & 0.89 & 0.52 &  m\\
  z2  & 0.32 & 0.19 &  m\\ 
  z3  & 0.36 & 0.22 &  m\\
  m1  & 45.7 & 19.6 & kg \\
  m2=m3  & 12.15 & 5.2 & kg \\
  $w/2$  & 0.1 & 0.06 & m \\
  \hline \\
  \end{tabular}
  \caption{Parameters of 3LP model for adult-size and kid-size models used for simulations in this paper. }
  \label{table::params}
\end{table}

\subsection{Pseudo-passive gaits manifold}
\begin{figure}[]
        \centering
        \includegraphics[trim = 0mm 0mm 0mm 3mm, clip, width=0.5\textwidth]{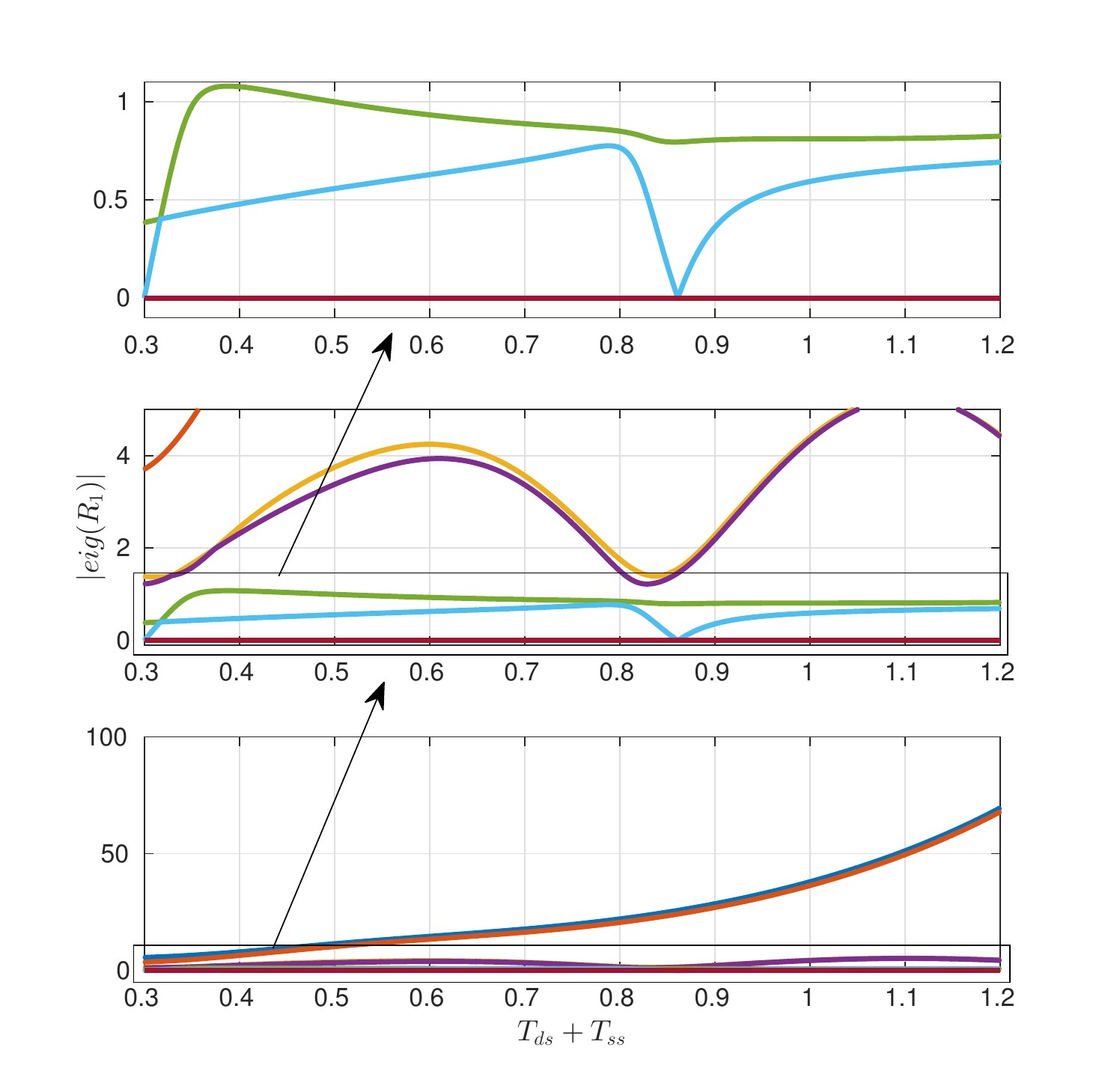}
        \caption{Square-root of the singular values of the matrix $R_1^TR_1$ with respect to the stride time $T_{stride}$, plotted for a adult-size model. In these plots, we have fixed the double support time $T_{ds}=0.3s$. It is notable that around $T_{stride}=0.86s$, the system shows a zero singular value which corresponds to a null-space containing infinite number of periodic solutions. These solutions are all without swing hip or stance contact actuation, referred to as pseudo-passive gaits.} 
        \label{fig::eig_passive}
\end{figure}

The first choice is to see whether the system has any pseudo-passive walking pattern or not. By pseudo-passive we mean a gait in which swing hip and stance contact torques are zero. The term pseudo is indeed indicating that in stance hip, the actuators are producing or dissipating power. It also refers to the fact that in our linear model, the legs are stretched or shortened by prismatic actuators, as part of model construction. To find pseudo-passive gaits, we look at the reduced matrix $R_1\in \mathbb{R}^{8\times 7}$ (not necessarily square) extracted from $R_0$ by excluding also the 8 columns attributed to hip/ankle torques. Since any valid solution should lie in the null-space of this matrix, we are interested in inspecting singular values. We normally calculate singular values of $R_1^TR_1$, since any vector in the null-space of this new matrix lies in the null-space of $R_1$ as well. The resulting singular values are shown in Figure.\ref{fig::eig_passive} over time.  

One can clearly see that there is a time $T_{stride} = T_{relax}=0.86s$ where the system shows two zero singular values. $T_{relax}$ can be found by a simple root-finding algorithm. It is obvious that one of these singular values refers to the sagittal and the other to lateral dynamics. We can simply calculate corresponding eigenvectors of $R_1^TR_1$ and find the null-space manifold, composed of $v_1$ and $v_2$. Note that there is only one lateral solution as the value of $d$ should only be $\pm 1$. However the solution in the sagittal plane can be scaled by any arbitrary positive or negative value to obtain different modulated speeds. Therefore the manifold of pseudo-passive compass gaits in this case is only 1-dimensional. From such analysis, we can also conclude that if any other stride time is chosen, the robot cannot demonstrate pseudo-passive forward progression and only steps in place. A demonstration of normal pseudo-passive compass gaits can be found in Figure.\ref{fig::snapshots}.

\subsection{Actuated gaits manifold}
\begin{figure}[]
        \centering
        \includegraphics[trim = 0mm 0mm 0mm 3mm, clip, width=0.5\textwidth]{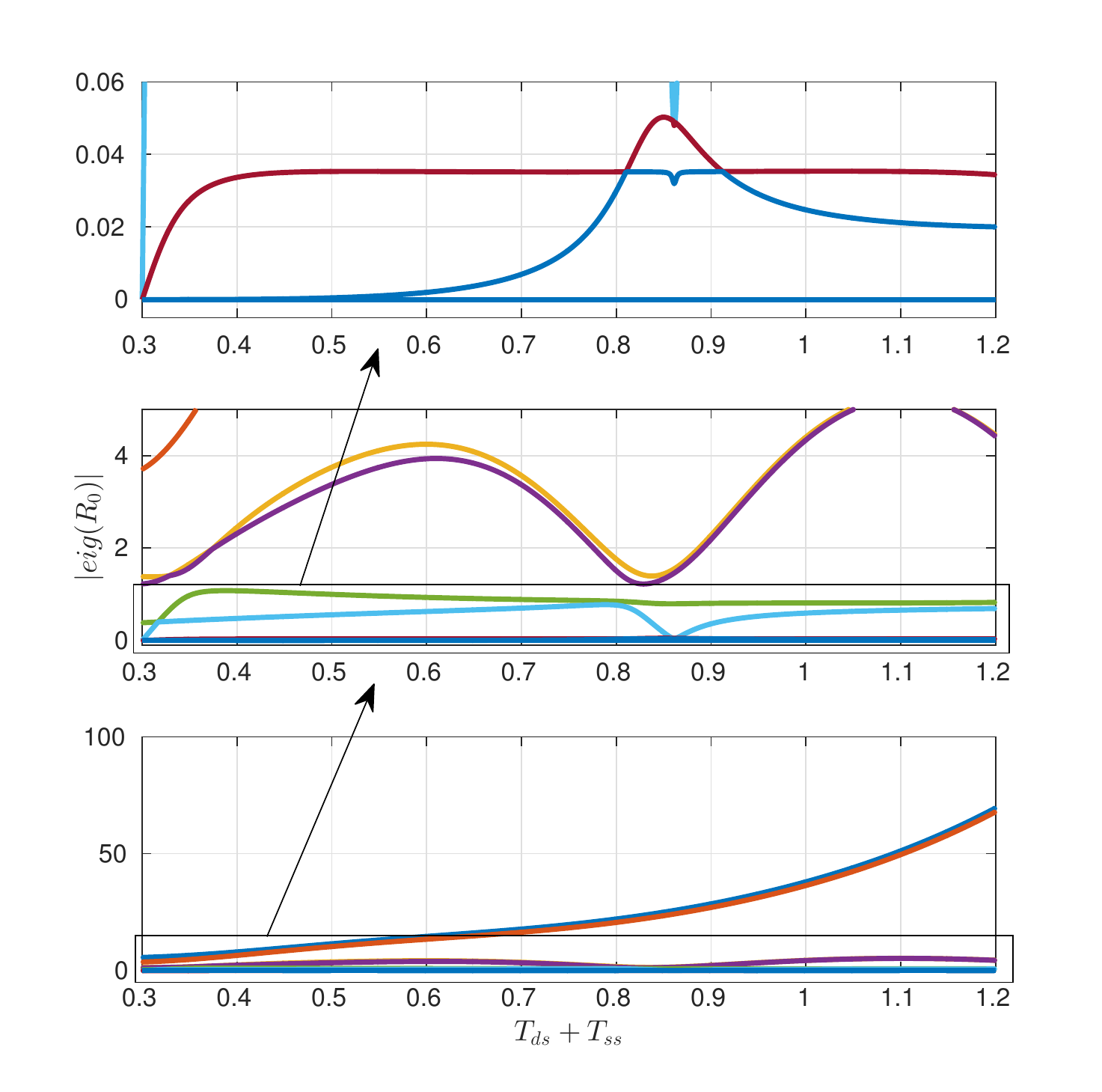}
        \caption{Square-root of the singular values of the matrix $R_0^TR_0$ with respect to the stride time $T_{stride}$, plotted for a adult-size model. In these plots, we have fixed the double support time $T_{ds}=0.3s$. It is notable that around $0.9s$, the system does not show a zero singular value like before. However there are 7 default zero eigenvalues that can produce gaits for any choice of $T_{stride}$. These gaits are indeed actuated, with many possibilities of swing hip or stance contact torque profiles.} 
        \label{fig::eig_hip}
\end{figure}

In this part we are going to find manifolds of motion which can benefit from swing hip actuation and CoP modulation as well. These inputs are of course containing a constant and a time varying components for both sagittal and lateral dynamics, as discussed in the previous section. With this ability, we can pump energy to the swing leg and brake at the end to produce faster swing motions. We can also apply contact torques which modulate the CoP and resemble the fact the CoP in human goes forward from the heel to the toes over a swing phase. We simply perform all calculations on $R_0$ itself. The resulting singular values using the same method described earlier are shown in Figure.\ref{fig::eig_hip} over time.

It is surprising that the system does not have a distinct zero singular value at $T_{relax}$ like before. However, it has 7 singular values equal to zero that produce a null-space at any given stride time. Each of the corresponding singular vectors $v_i$ where $1 \le i \le 7$ have similar dimensions with $Q$ in (\ref{eqn::full_vector}), though with $P$, $\dot{X}_{2}$ and $W$ to be zero. These are nominal initial states with contact point at origin, resting swing foot and no disturbance of course. This null-space is not 7-dimensional however. The variable $d$ which should always be $\pm 1$ reduces the total dimensions to 6 like before. Now one can simply decide the timing and active actuators in order to reduce this high dimensionality and find a unique solution. 

Now for any desired speed, we have the possibility to choose $T_{ds}$, $T_{ss}$ and a linear combination of resulting 7 null-space vectors. As demonstrated in pseudo-passive gaits, at $T_{relax}$, a certain combination of these vectors can results in zero-torque gaits. Now what if we calculate the 7-dimensional null-space using $T_{relax}$? Could we still find a pseudo-passive linear combination? In fact it is possible, although no distinct zero singular value is observed in Figure.\ref{fig::eig_hip}. The reason is that the rank of actuation space at $T_{relax}$ is equal to 5 in the 7-dimensional null-space manifold. This means if we constrain all of them to zero for pseudo-passive walking, we only loose 5 ranks. The other 2 ranks are therefore dedicated to the variable $d$ and the desired speed, like before. So the null-space manifold of actuated gaits already encompasses the one for pseudo-passive gaits and we do not need to calculate them separately. In the next subsection, we are going to show a few examples of walking gaits using the null-space calculated.

\subsection{Numerical examples} 
Apart from pseudo-passive walking which has a certain timing $T_{relax}$, we are going to show same speed walking solutions with different timing, once using only hip and once using hip and ankle torques. From singular value analysis, we have 7 singular vectors that can produce motion. We pack them together column-wise in a matrix $V=\begin{bmatrix} v_1 & v_2 & ... & v_7 \end{bmatrix}$. We also select a more human-like choice of $T_{ds}=0.1s$ and  $T_{ss}=0.6028s$ calculated from pseudo-passive walking. The choice of speed will be $v_{des}=1m/s$. 

\begin{figure}[]
        \centering
        \includegraphics[trim = 20mm 10mm 20mm 10mm, clip, width=0.5\textwidth]{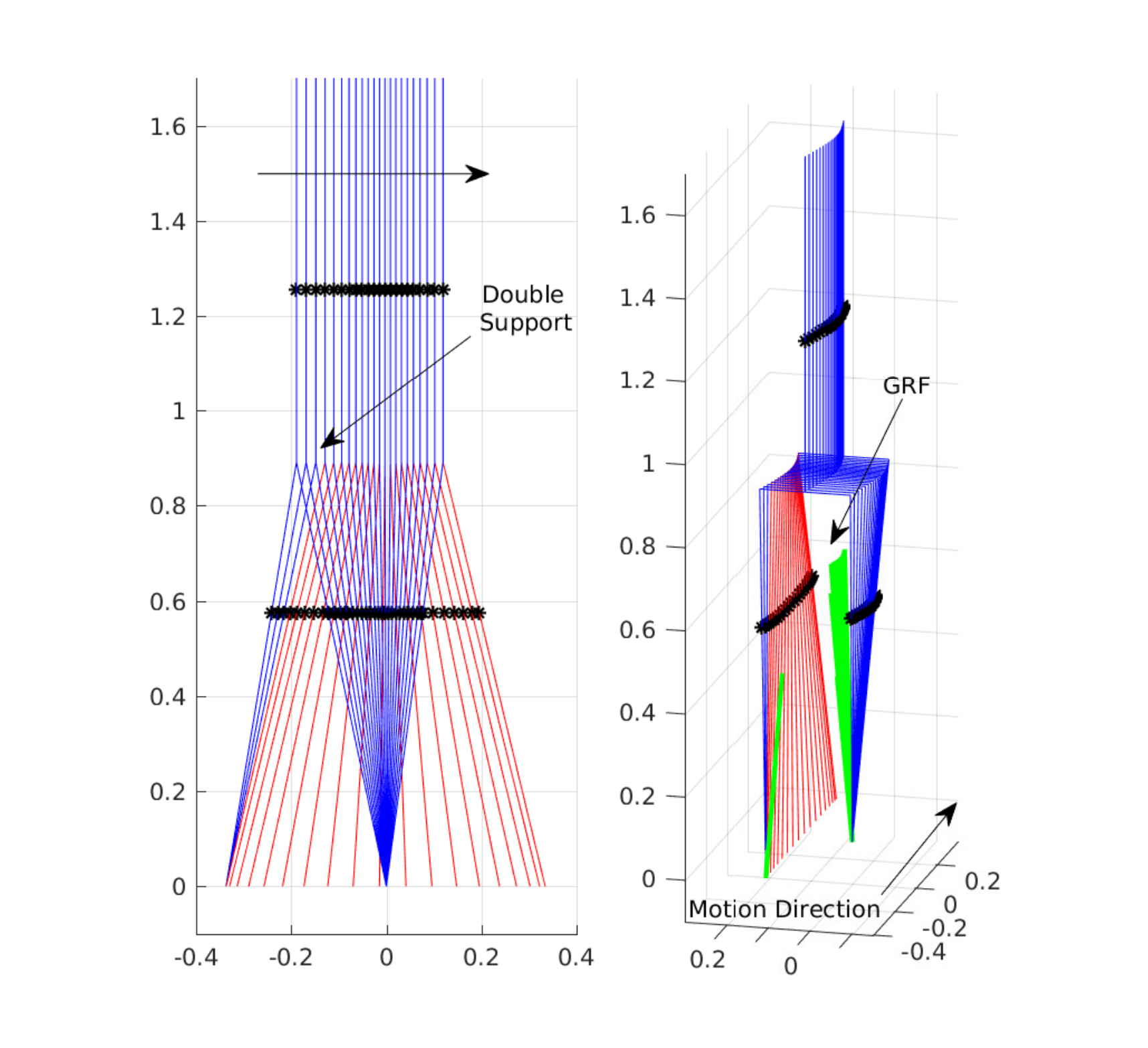}
        \caption{A detailed demonstration of a full stride phase in pseudo-passive walking where snapshots are taken every $30ms$. Black arrows show the direction of motion and the swing leg is shown in red. In this figure, lateral bounces could be seen on right while velocities can be inferred from the snapshots on left. The swing leg speeds up and slows down during a stride phase while the torso has minimal speed when the swing foot is at maximum speed. It can also be observed that the swing foot approximately follows a straight line while the swing hip bounces laterally. } 
        \label{fig::snapshots}
\end{figure}

\begin{figure*}[]
        \centering
        \includegraphics[trim = 50mm 10mm 40mm 0mm, clip, width=1\textwidth]{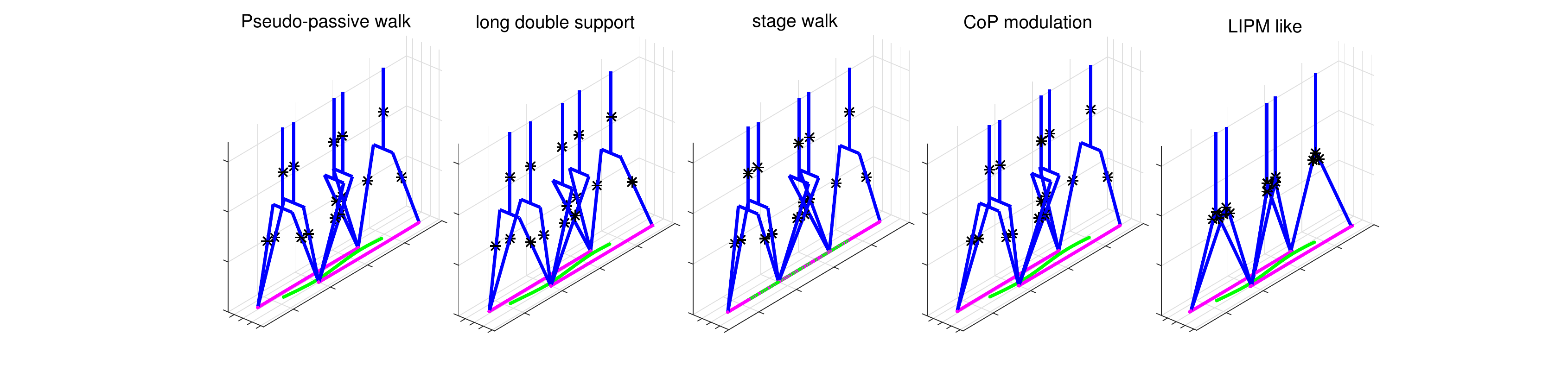}
        \caption{Snapshots of different walking scenarios at $1m/s$ and $1.5 step/s$. These snapshots are taken at the switching times from double support to single support or vice versa. Feet trajectories are plotted along with the projection of the CoM trajectory on the ground. In pseudo-passive walking, there is no actuation. However one can clearly see that the model is able to produce CoM trajectory, lateral bounces and swing dynamics. In long double support case, the motion is geometrically quite similar. Stage walking is basically producing no lateral bounce and uses proper hip torques to let the model step only on a single straight line. On the real robot however, one should avoid foot scuffing and this motion is not feasible. CoP modulation also shows quite similar geometry to pseudo-passive walking, though CoM trajectory (the green line on the ground) starts a bit further from the trailing leg, producing a more symmetrical gait. The influence of CoP modulation therefore is mainly on sagittal symmetry and less variations in the sagittal speed (Figure.\ref{fig::vCoM}). Finally, the motion of LIP like model is also rather similar to pseudo-passive case, but we will see that it has higher CoM speed variations (Figure.\ref{fig::vCoM}). In this case, we impose lateral footstep distances to be similar to other scenarios. Otherwise, in LIPM pelvis width is not modeled. All corresponding walking movies could be found in Multimedia Extension.} 
        \label{fig::walk}
\end{figure*}

We setup an optimization problem to find linear coefficients $\alpha_i$ which produce walking gaits of desired speed $v_{des}$ with minimal input torques. As mentioned before, this is just a demonstration and does not have any precise meaning in terms of energy. The purpose of this paper is not to find a cost function for energy that produce human-like motions. It is part of our future work to compare this model with human gaits and their associated timing. In \cite{asano2004novel, hasaneini2013dynamic} however, limit cycles for the real robot are found through dynamic energy minimization which is interesting and inspiring for future works. By considering a vector of $\alpha=\begin{bmatrix} \alpha_1 & \alpha_2 & ... & \alpha_7 \end{bmatrix}^T$, our optimization is formulated as:
\begin{eqnarray}
\begin{aligned}
& \underset{\alpha}{\text{minimize}}
& & |S_{U}V\alpha|^2 \\
& \text{subject to}
& & S_{d}V\alpha = \pm-1 \\
&&& S_{X_{2,x}}V\alpha = -v_{des}(T_{stride})
\end{aligned}
\end{eqnarray}
Now consider the following scenarios:
\begin{itemize}
	\item \textbf{Pseudo-passive walk:} which is being calculated as mentioned before. Note that it can be shown if $T_{relax}$ is chosen, the result of our optimization is the same pseudo-passive gait, as obtained before. 
	\item \textbf{Long double support:} in this case we enforce ankle torques to zero by adding another constraint to the optimization:
	\begin{eqnarray}
		S_{M_{3}}V\alpha = 0
	\end{eqnarray}
	Keeping the same stride time, we double $T_{ds}$ and decrease $T_{ss}$ accordingly. Note that now the walking cannot be pseudo-passive anymore, so the optimization will find minimal hip torques to produce the same speed and stride length. 
	\item \textbf{Stage walk:} here we constrain ankle torques to zero like before. Now instead of optimizing torques, we optimize the lateral velocity in the cost function. The optimization will give hip torques that produce a motion with minimal lateral bounce. In this case, the biped walks on a straight line without lateral bounce.
	\item \textbf{CoP modulated:} given the length of human foot, the total weight and the timing of single support, we can calculate a constantly increasing ankle torque $\frac{t}{T_{ss}} \tau_{CoP}$ acting in sagittal plane to move the CoP to the toes gradually during single support. In this scenario, we force other components of ankle torques to zero and the time-increasing component to $\tau_{CoP}$ by adding the following constraint to the optimization:
	\begin{eqnarray}
		\begin{bmatrix} S_{M_{a}} \\ S_{rM_{a}} \end{bmatrix}V\alpha = \begin{bmatrix}
		0 & 0 & \tau_{CoP} & 0
		\end{bmatrix}^T
	\end{eqnarray}
	The result is a gait with time-increasing ankle torque profile and proper hip actuation, walking at the same frequency and speed like before.	
	\item \textbf{LIP like:} in this case, keeping the original timing, we change the model of the robot. We move most of the weight of each leg to the torso, and also move the 3 masses closer to the pelvis by decreasing $z2$ and $z3$. The idea is to see a behavior similar to LIP. Again we disable all ankle torques as well.  
\end{itemize}
It should be noted that these optimizations are very fast, in the order of microseconds. It is also always possible to find closed-form solutions as the optimizations are equality-constrained quadratic optimizations. The accompanied Multimedia Extension demonstrates different features of 3LP while Multimedia Extension shows movies of five previously mentioned scenarios. The 3D geometry of resulting gaits are shown in Figure.\ref{fig::walk} while a detailed diagram of each stride is shown in Figure.\ref{fig::snapshots}. Our flexible model can produce periodic gaits with different actuation schemes. We consider piecewise linear profiles for each actuator to produce more human-like torque and ground reaction profiles, investigated in the final section. 

\begin{figure}[]
        \centering
        \includegraphics[trim = 0mm 0mm 0mm 0mm, clip, width=0.5\textwidth]{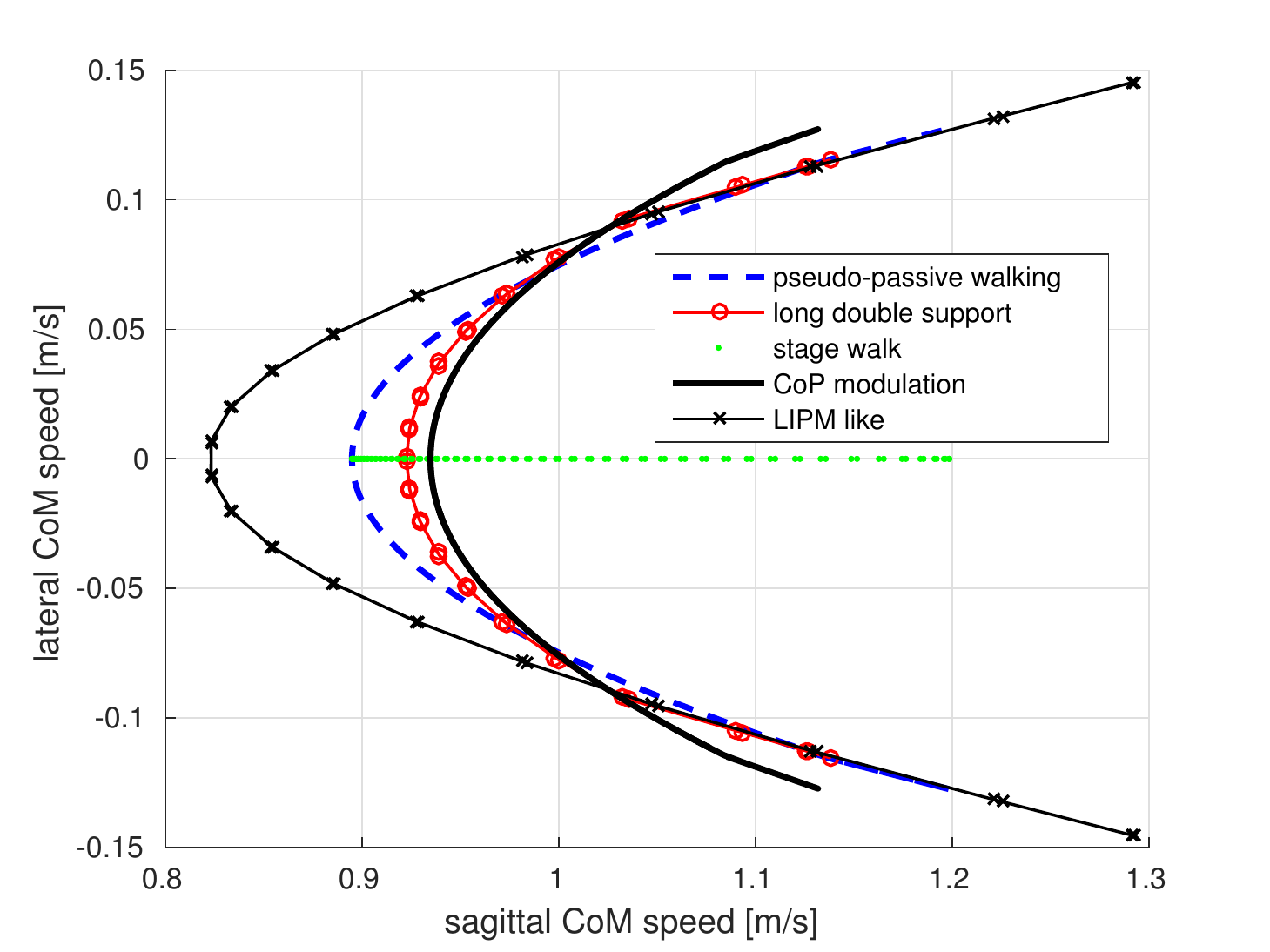}
        \caption{Sagittal vs lateral CoM velocity trajectories for different scenarios discussed. Note that LIP like model produces large sagittal variations. Pseudo-passive walking is still showing high variations in both directions. Larger lateral motions in pseudo-passive walking with respect to LIP might be explained by higher altitude of CoM. It can be concluded however that swing and torso dynamics clearly reduce these variations. Long double support also reduces variations in both directions. By modulating CoP, although lateral motion is the same as pseudo-passive walking, sagittal variations are reduced even more and the motion is smoother. Finally, one can see that stage walking has no lateral motion, however it has similar sagittal motion to pseudo-passive walking. In general, we can conclude that increasing double support time has similar effect on CoM speed variations as CoP modulation. However it does not induce any argument on energy efficiency.} 
        \label{fig::vCoM}
\end{figure}

It can be concluded from Figure.\ref{fig::walk} that changing different parameters does not have major effect on the overall geometry of walking. However, since our target is to develop a template model that is easy for inverse dynamics to track, we are interested to investigate dynamic properties of these walking scenarios. For this purpose, we have shown CoM velocities for all scenarios in Figure.\ref{fig::vCoM}. 

Although CoM trajectories look similar in Figure.\ref{fig::walk}, they have very different characteristics in terms of velocity variations. The LIP like model shows a large variation in sagittal velocity. It is not so obvious how swing and torso dynamics affect this motion at first glance. Remember that by torso dynamics, we mean the torque required by the stance hip to keep the torso always upright. This torque is not necessarily zero, since the pelvis is an accelerated frame. Therefore, hip torque can affect CoM motion considerably, specially since the torso is relatively heavy. Moreover, although the swing foot has smaller weight compared to other parts of the body, it can be seen from Figure.\ref{fig::walk} that swing leg has a faster motion during single support. Such fast motion quadratically increases kinetic energy and therefore results in a considerable work flow. In our model, we have described these effects in a simplified and linear fashion, but capturing important couplings between the 3 pendulums. 

Taking a closer look at Figure.\ref{fig::vCoM} reveals that even maximal CoP modulation still does not change velocity profile considerably. This basically means the difference between pseudo-passive walking and LIP is way larger than that between pseudo-passive and CoP modulated walking. In other words, CoP authority can at most convert the pseudo-passive gait to the CoP modulated gait. The available CoP authority is hardly enough to convert the pseudo-passive gait to LIP gait and this gap increases mainly in faster walking speeds. Although here the speed is moderate, we can easily infer that LIP as a template model can only operate in a very limited range of walking speeds. Remember that in fact an inverse dynamics or kinematics approach eventually synthesizes the template motion with the full model by exploiting all control authorities of the robot (including CoP). This motivates therefore not to modulate the CoP in template level and leave the control authority free for full-body controllers to mimic the template motion as precisely as possible. 

In this section we discussed an easy method to find manifolds of periodic motions without any numerical forward simulation of the system. Once these manifolds were found, we also showed how to find individual solutions, based on the type of actuation and timing desired. We only considered gaits with minimal hip torques here. However to go further, we would like to investigate the effect of timing and walking speed as well. Such investigation reveals interesting energetic properties of 3LP, discussed in the next section.

\section{Comparison with human data}

\begin{figure*}[]
        \centering
        \includegraphics[trim = 20mm 5mm 20mm 5mm, clip, width=1\textwidth]{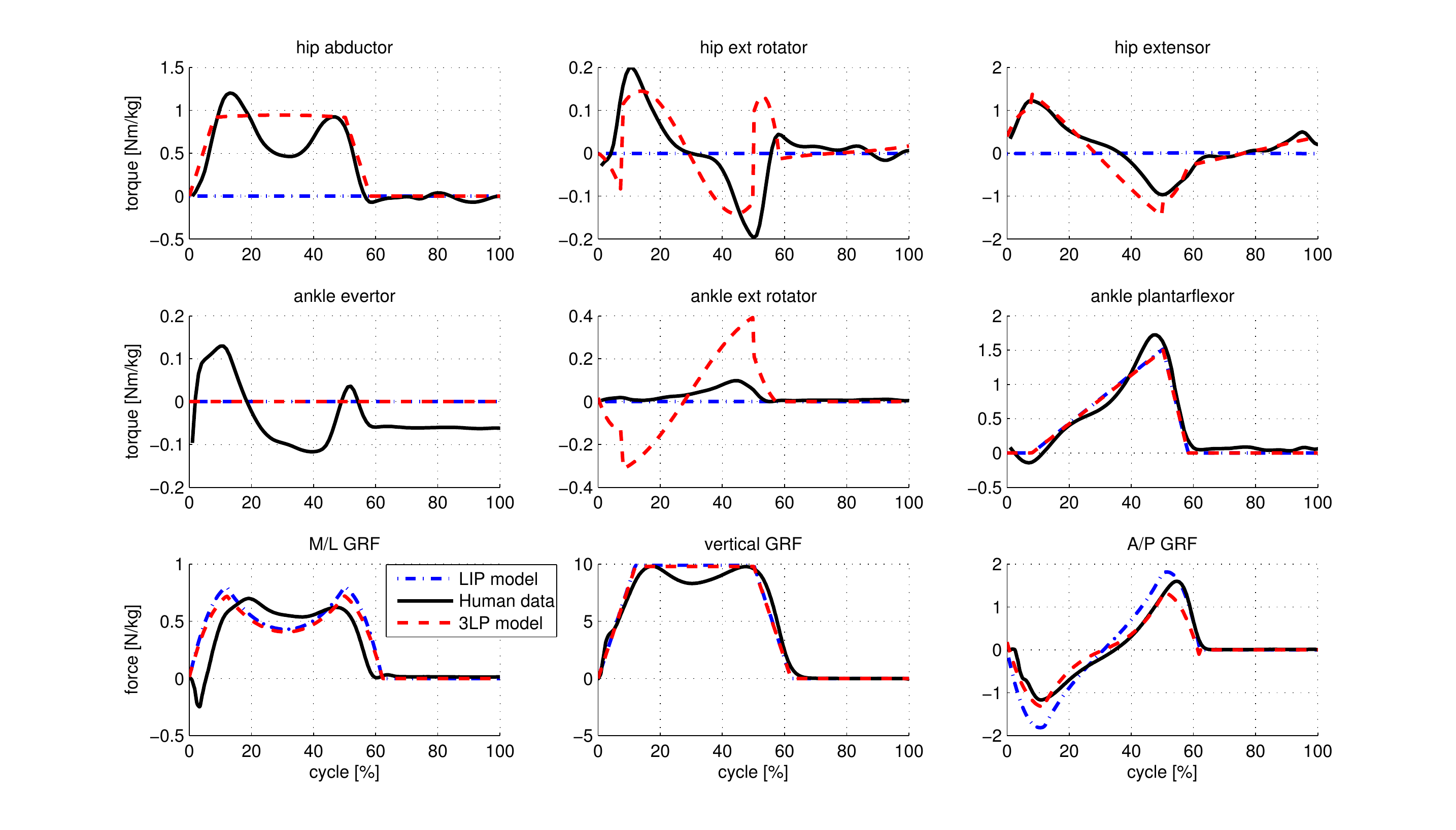}
        \caption{Comparing dynamic profiles of 3LP and LIP with normalized human data, taken from \cite{eng1995kinetic}. This data is for male subjects with average weight of $77.2kg$ and height of $1.8m$, walking at $1.6m/s$ and $108steps/min$. In these curves, we have demonstrated hip/ankle torques as well as ground reaction forces. Note that our model does not have any knee and we consider ankle torques to be approximately equal to contact wrenches. Although all profiles of 3LP match the human data quite well, lateral and transversal dynamics have some discrepancies. The LIP model is however unable to describe hip torques as it does not include swing and torso dynamics. Here we use the same CoP modulation for both LIP and 3LP for better comparison.} 
        \label{fig::profiles}
\end{figure*}

Compared to LIP, the 3LP model is much more similar to human locomotion, because it describes falling dynamics, swing motion, torso balance, lateral stepping and double support features all together. In addition to geometric similarities, we plot ground reaction forces and joint torques to compare the underlying dynamics that result in such geometric similarity. For this purpose, regarding the available data from human subjects \cite{eng1995kinetic}, we select similar model parameters and timing, calculate periodic manifolds and find solutions with the same speed and CoP modulation pattern. Such comparison is demonstrated in Figure.\ref{fig::profiles}. In the following, we discuss various similarities observed in this figure. 

\subsection{Sagittal dynamics} 
From the last column of Figure.\ref{fig::profiles}, one can observe a good match of hip extensor and ankle plantarflexor torques as well as Anterior-Posterior ground reaction forces. This is despite the fact that walking speed is relatively fast and the step length is about $80\%$ of the leg length. Note also that the constant and time-increasing components for hip/ankle torques are roughly enough to describe major trends in human curves. Nonlinear profiles of 3LP are however related to the stance leg and generally those degrees of freedom which are not directly controlled by desirable input torques. The LIP model does not have hip torques and produces larger A/P GRF, because of different CoM trajectories shown in Figure.\ref{fig::vCoM}.

\subsection{Vertical GRF} 
By model design, the CoM height is constant and we do not expect two peaks in the model profiles, similar to \cite{rummel2010stable,maziar2015FMHC}. But the general trapezoidal shape is preserved, thanks to our double support phase. The main consequence of such profile is walking with crouched knees which looks less human-like compared to other template models. Note that LIP model produces the same profile as expected.

\subsection{Transversal rotation torques} 
The transversal rotation torques are preserving the general trend of the human data, but not matching very well, specially in the ankle. One major reason is that arm motions and pelvic rotations are not considered in the model. The other important reason is that here we just minimized all hip/ankle torques to find a unique solution out of the manifold of all available solutions. This minimization is not necessarily realistic and human-like, as it causes wider lateral stepping, larger lateral motion and therefore more ground moment. A better cost function on energy might produce more human-like gaits, although it is doubted in \cite{workman1986metabolic} that optimal gaits are defined merely by energy terms. They might be highly influenced by posture balance, at least in lower speeds. In faster walking speeds like the human data demonstrated here \cite{eng1995kinetic}, humans tend to take laterally closer steps, compared to our model. Note that the transversal moment is required to keep the torso upright and straight ahead during the swing phase, compensating the moment produced by the swing leg. In LIP however, since there is no pelvis and inertia around the yaw axis, we do not expect transversal torques. 

\subsection{Lateral dynamics} 
Again due to previously mentioned problem in optimizing lateral motion, we see that our model keeps general trends, like double peaks in the Medio-Lateral GRF, but cannot precisely describe other torque profiles. Humans normally tend to step as close as possible to minimize lateral motions and energy \cite{kuo1999stabilization} (remember stage walking in Figure.\ref{fig::walk}). However humans swing their foot over an arc shape to avoid scuffing as well. Such fine motion requires better objective functions to find proper solutions from the available manifolds. Note that the sagittal swing motion can influence lateral dynamics as well \cite{kuo1999stabilization, collins2005efficient}, possibly through transversal moments. This might be another reason for the discrepancy observed between different lateral curves. Our model however completely decouples lateral and sagittal dynamics. The only linking parameter is stride time which relates to the zero-velocity assumption for the swing foot. In the LIP model, there is no pelvis modeled. However we consider a gait with the same lateral foot-stepping for better comparison. We can observe that although LIP does not explain hip torques, M/D GRF forces are yet similar to 3LP.

\subsection{Energetics}
Apart from dynamics profiles, it is always interesting to investigate energy flow in the model and compare it with the real human. Remember that one of the main motivations behind developing 3LP is to match humanoid dynamics as precise as possible in the template space. Such matching could be viewed form the power-flow perspective, inspired by the fact that humans can walk very efficiently. Compared to LIP, 3LP can additionally describe swing and torso dynamics which are important aspects of locomotion, specially in faster speeds. Since the pelvis has accelerations in the sagittal and lateral directions, the whole upper-body requires a hip torque to remain upright during locomotion. Our model considers stance hip torques to fulfill this requirement and of course describes the influence of these torques on horizontal motions. We do not model torso movements in 3LP and assume they are negligible, but they might induce additional hip torques too. Although the weight of swing leg is relatively small, its peak velocity is about two times larger than the torso which has a heavier mass. Therefore, the peak kinetic energy of the swing leg is quite comparable with the torso. Such energy comes from both accelerated pelvis where the swing leg is attached to and the swing hip torques that can change swing dynamics. In trade off with the work required for accelerations and decelerations of the torso, this phenomenon explains optimal speed-frequency relations observed in metabolic cost of human walking \cite{bertram2005constrained}. 

Although 3LP does not describe many important walking features like heel-toe motions, knee flexions and CoM excursions, it is still very surprising that it can capture the main optimality trend in human walking. To demonstrate this capability, we consider two main parameters of locomotion, stepping frequency and forward speed. Inspired by \cite{bertram2005constrained}, we calculate the economy of walking for different speed-frequency combinations. Such Economy is in fact the inverse of cost of transport, which is the total energy consumed per unit mass, per unit distance traveled. There are many ways to calculate such mechanical power in our model. In fact, 3LP does not simulate muscles and exact geometry of a human and is thus, unable to precisely reconstruct the human economy surface based on the metabolic cost. However by integrating the mechanical power on CoM, we are able to approximate a portion of this energy which still plays an important role in the overall energy, according to \cite{anderson2001dynamic}. We are not able to model other costs like muscle activation, maintenance and shortening heat rates though, since we do not have muscles in 3LP.

\begin{figure}[]
        \centering
        \includegraphics[trim = 0mm 25mm 0mm 15mm, clip, width=0.5\textwidth]{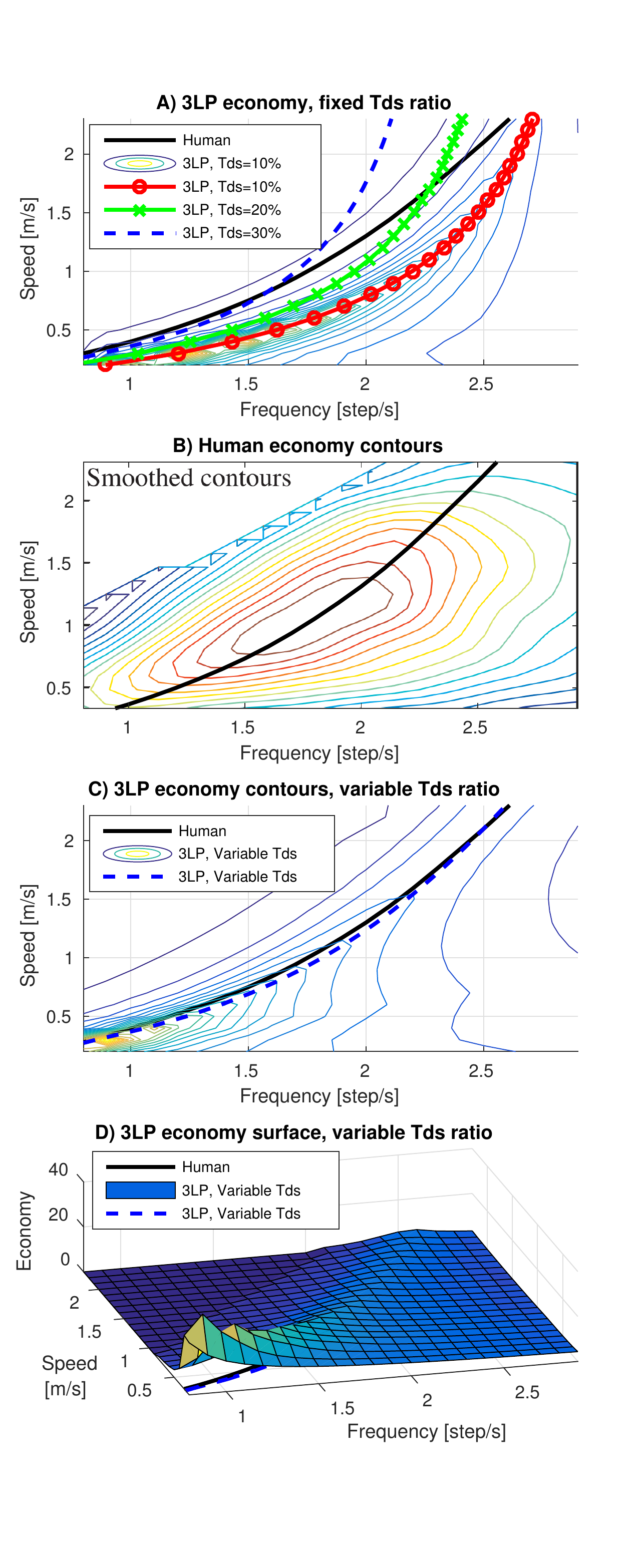}
        \caption{Economy of walking, i.e. the inverse of Cost of Transport (CoT) calculated as positive work on CoM over unit distance traveled, normalized by body mass. Here we use the body mass of $66kg$ and height of $1.7m$ with normal human-like mass distribution. \textbf{A)} Contours of 3LP walking economy, calculated for the choice of $\frac{T_{ds}}{T_{stride}}=10\%$. Optimal peaks for this ratio as well as other choices of $20\%$ and $30\%$ are plotted versus optimal relation in human. Here 3LP gives an optimal peak, but not matching with the human data. \textbf{B)} The smoothed 3D surface of human walking economy, taken from \cite{bertram2005constrained}. \textbf{C)} 3LP Economy, using the natural choice of (\ref{eqn::optimal_tds}), obtained from human data. Now, 3LP almost precisely shows similar trend with human. \textbf{C)} The actual 3D surface of 3LP economy, associated with the plot C. } 
        \label{fig::energy}
\end{figure}

By doing a systematic search over a grid of different speed-frequency combinations, we calculate the net positive work performed on the CoM per unit distance, divided by the total mass. The calculation of such energy is rather easy in our model, as it corresponds to the difference between minimum and maximum kinetic energy. Since the velocity of the CoM is a linear function of the state variable, it can be simply related to the known initial state through $H(t)$ matrix which is a combination of exponential functions of time. (\ref{eqn::Hss_decompose},\ref{eqn::Hds_decompose}). The problem reduces to solving two simple maximization and minimization problems which are fast, thanks to the simplicity of closed form solutions. Figure.\ref{fig::energy}.A demonstrates economy contours for the choice of $\frac{T_{ds}}{T_{stride}}=10\%$. It is surprising that the model is showing a peak line within the desired range of frequency-speeds. Such peak in fact demonstrates the trade-off between hip torques and contact switching acceleration/deceleration costs. The former increases in higher frequencies as energy needs to be pumped into the swing leg and taken out by braking at the end of swing \cite{doke2005mechanics}. The latter however increases by step-length which results more variations in the CoM velocity. 

We have also repeated the same process for two other double support time choices of $20\%$ and $30\%$. The resulting peak lines are demonstrated on Figure.\ref{fig::energy}.A as well as the optimal trend of human data, taken from \cite{bertram2005constrained} (the case of constrained speed). Here we use the same average weight ($66kg $) and height ($1.7m$) of human subjects in \cite{bertram2005constrained} as well as average weight distribution reported in \cite{de1996adjustments}. Although 3LP is successful in identifying the trade-off, using the choice of constant double support time ratio, it fails to match the human data. 

It can be postulated that in slow speeds, humans have larger double support time ratios compared to higher speeds (refer to Figure.\ref{fig::energy}.A). Experiments on real humans actually verify this postulation, suggesting a linear relation between double support time and the walking speed \cite{cappellini2006motor}. Here we implement similar simple law, making $\frac{T_{ds}}{T_{stride}}$ a linear function of walking velocity $v$:
\begin{eqnarray}
	\frac{T_{ds}}{T_{stride}} = 0.12 + (2.5-v)\times0.09
	\label{eqn::optimal_tds}
\end{eqnarray} 

This relation gives a ratio of $27\%$ at $v=0.8m/s$ and $12\%$ at $v=2.5m/s$, close to our three initial conjectures. Repeating the same search process with this particular choice of double support time, we obtain Figure.\ref{fig::energy}.C with the actual economy surface shown in Figure.\ref{fig::energy}.D. It is surprising that 3LP is able to predict almost precisely the energy-optimal relation between frequency and speed in human walking. Although 3LP itself cannot probably explain the optimal double support time, it already shows that minimal knowledge of human data is enough. The dependence of (\ref{eqn::optimal_tds}) on body weight and geometry is yet to be explored in future works. If not dependent, the equation (\ref{eqn::optimal_tds}) and 3LP seem to be enough to predict optimal frequency-speed relation, given specific body parameters. 

Figure.\ref{fig::energy} is very promising and important, despite the fact that:
\begin{itemize}
	\item No impact or push-off is considered.
	\item CoM height is constant.
	\item No foot clearance is modeled.
	\item Heel-toe motions and knee flexion are not included. 
	\item Upper body is assumed to be fixed.
	\item The torso has no rotation in any direction. 
	\item Weight loading cost on stance leg is not considered.
	\item Walking-independent metabolic cost is not modeled.
\end{itemize}
Therefore, although the overall frequency-speed relation is correctly predicted, the 3D surface of economy is not matching the human data shown in Figure.\ref{fig::energy}.B, taken from \cite{bertram2005constrained}. The correct prediction in high-frequency and high-speed walking conditions is surprising. Linear models are conventionally expected to be a linearization of the actual non-linear system, thus performing well in near-stance (small stride-length) conditions where the geometry of 3LP is more close to the actual system. However here, although 3LP does not model heal-toe-knee motions and intrinsically simplifies them with an extensible prismatic actuator, it appears to demonstrate the same mechanical energy flow, even with large stride length. In future works, it is still interesting to explore other sub-components of walking energy, keeping 3LP as a core model. Inclusion of other costs might reconstruct the actual human economy surface more precisely.

Overall, compared to LIP, while being still linear and slightly more complicated, the proposed model is able to describe much more features of human walking. It is very important in hierarchical control architectures that template models produce as accurate motions as possible to make tracking more precise. Despite being linear and of course operating in a more limited region of feasible states, this model is better than LIP to describe human dynamics and consequently, more precise for controlling humanoids. The energy flow of 3LP is also more similar to human than LIP, making the planned motion more natural for the humanoid robot which has similar body features.

\section{Conclusion}

Compared to most of other template models listed in Figure.\ref{fig::models}, our proposed model considers swing and torso dynamics in a linear formulation. On the other hand, it is computationally similar to LIP which is vastly used in the literature to control real robots over relatively slow walking speeds \cite{sakagami2002intelligent}. Other nonlinear models are also used in simpler robots \cite{collins2005efficient}, but over a limited range of speeds. Template models try to describe major dynamics of the robot in an abstract way. This can be used either for analysis of human motion or control of a complex robot, probably in a hierarchy with more complex controllers. In such control paradigm, it is important to keep computational costs as minimal as possible, favoring future prediction. 

In 3LP model, the pelvis width links linearly with the lateral motion. This parameter can be used to find lateral ankle/hip torques required for balancing and to determine natural lateral foot placement. This is compared to most of other methods like \cite{faraji2014robust} where the two feet are enforced to be apart to avoid scuffing. In literature, the timing and footstep locations are mainly enforced by other desired trajectories. In our model however since swing dynamics is included, we introduce zero foot velocity assumption and let natural periodic gaits come out of equations. This assumption which relieves the need to calculate impact forces actually determines the timing and periodicity conditions as well.

The proposed model can predict human walking profiles quite well, even for relatively fast speeds where the linearity assumption might be quite limiting for operation on the real robot. Although IP-based models can demonstrate CoM excursions quite well, their nonlinear nature makes them less suitable for highly complex robots that require online planning. There are more advanced versions of IP-based models, including torso and swing dynamics. However again, nonlinear equations cannot be used for per time-step future prediction over a wide range of speeds, although they might better describe energetics of human walking. The proposed model keeps a trade-off between geometric and dynamic similarities, favoring fast computation properties. 

We also showed that 3LP can approximately describe the exchange of energy despite keeping the CoM height constant which is a less realistic assumption in fast speeds. We would like to mention that the vertical excursion of CoM in human locomotion, even in very fast walking at $2m/s$ is still about $5cm$ \cite{gard2004comparison} which is quite negligible compared to the leg length of about $1m$ (pelvis excursion is about $7cm$ however). CoM excursion is mainly determined by the stride length which is not much larger in faster speeds. Humans in fact increase stepping frequency together with step length to walk faster, which highly affects swing dynamics and requires more energy pumped by hip torques. Therefore in faster speeds, apart from the double-peaked vertical GRF, one should consider swing dynamics as well due to considerable exchange of energy. LIP of course fails to describe swing dynamics, but 3LP successfully captures them. 

We have not used it for extensive bio-mechanical analysis in this paper. Rather, we focus on the fact that such similarity can be inspiring for generating more precise abstract plans, used to control humanoid robots. In brief, advantages of the 3LP can be listed as following:
\begin{description*}
	\item[+] Swing dynamics.
	\item[+] Torso balancing.
	\item[+] Double support.
	\item[+] Optimal frequency-speed relation similar to human.
	\item[+] Hip/ankle actuation possibilities.
	\item[+] Computationally fast.
	\item[+] Possibility to consider hip torque limits.
	\item[+] Natural lateral motion.
	\item[+] Natural periodic gaits.
	\item[+] Pseudo-passive compass gait.
\end{description*}
While disadvantages are:
\begin{description*}
	\item[-] Flat vertical GRF profiles.
	\item[-] Stretching legs.
	\item[-] No steering capability yet.
	\item[-] No arm motion.
	\item[-] No torso pitch/roll DoF.
\end{description*}

It should be noted that without pelvis width like LIP, steering is possible as demonstrated in our previous work \cite{faraji2014robust}. Steering makes 3LP nonlinear, but one can compromise the lateral motion and let inverse dynamics find proper actuation patterns to turn around the yaw axis. In future works, we are going to replace the LIP with the proposed model and design better controllers to improve the performance on a full humanoid robot. We would also like to setup MPC with all aforementioned policies and inspirations to produce feasible plans for faster walking. 3LP can be extended to have two more degrees of freedom for the torso to describe asymmetries observed in human over fast walking speeds. It can also include actuation inputs proportional to the square of time to produce more precise torque profiles. Among many different advantages of 3LP, we favor its capability to produce more natural motions. In this way, we can expect our inverse dynamics layer to track the template model more precisely and therefore, being able to produce more human-like motions. This paper is accompanied with a multimedia extension, demonstrating general features of 3LP and the different gaits it can produce. All the codes used in this article as well as the multimedia extension are available online at \url{http://biorob.epfl.ch/page-99800-en.html}.

\section{Acknowledgments}
This work was funded by the WALK-MAN project (European Community's 7th Framework Programme: FP7-ICT 611832).

%
%
%

\bibliographystyle{IEEEtran}
\bibliography{Biblio}

\end{document}